\documentclass[11pt,twocolumn]{article}
\usepackage{pslatex, graphicx, amssymb, amsmath}
\usepackage{float}
\usepackage[section]{placeins}
\usepackage{booktabs}
\usepackage{tabularx}
\usepackage{multirow}
\usepackage{multicol}
\def\abstract{
\typeout{Abstract}
 {\bf Abstract} 
} 

\begin{document}
\title{SatSOM: Saturation Self-Organizing Maps for Continual Learning}

\author{Igor Urbanik\footnote{Corresponding author: iurbanik@student.agh.edu.pl}, Paweł Gajewski \\ Faculty of Computer Science, AGH University of Krakow,\\ al. Adama Mickiewicza 30, Krak\'{o}w, \ 30-059, Poland}

\maketitle

\begin{abstract}

Continual learning poses a fundamental challenge for neural systems, which typically suffer from catastrophic forgetting when exposed to sequential tasks. Self-Organizing Maps (SOMs), despite their inherent interpretability and efficiency, also exhibit this vulnerability. In this paper, we introduce Saturation Self-Organizing Maps (SatSOM)—an extension designed to enhance knowledge retention in continual learning scenarios. SatSOM incorporates a novel saturation mechanism that progressively reduces the learning rate and neighborhood radius of neurons as they accumulate information. This dynamic effectively stabilizes well-trained neurons, redirecting new learning to underutilized regions of the map. To further accommodate tasks of unknown complexity, we introduce a dynamic variant capable of adaptive grid expansion. We evaluate SatSOM on sequential versions of the FashionMNIST and KMNIST datasets, showing that it significantly outperforms existing SOM-based methods and approaches the retention capabilities of a k-nearest neighbors (kNN) baseline. Ablation studies confirm the critical role of the saturation mechanism. SatSOM offers a lightweight and interpretable solution for sequential learning and provides a foundation for implementing adaptive plasticity in complex architectures.

\end{abstract}

\section{Introduction}\label{sec1}

Intelligent agents functioning in real-world environments must continuously learn, adapting to new information while retaining prior knowledge~\cite{continual}. This ability, known as continual or lifelong learning, poses a significant challenge in modern machine learning. Most artificial neural systems struggle with catastrophic forgetting~\cite{catastrophic}, where training on new tasks or data distributions abruptly erases previously learned information. This phenomenon stems from the shared nature of representations in standard neural networks, where updating weights for new data can overwrite information critical for past tasks.

To quantify the extent of forgetting, we consider the k-Nearest Neighbors (kNN) algorithm as a useful reference point. Because kNN stores the entire history of training data and performs inference via direct comparison, it avoids forgetting by design. However, this retention comes at the cost of unbounded memory requirements and computational inefficiency during inference, making it impractical for scalable, real-world applications. In contrast, standard parametric models operate with fixed memory and computational budgets but generally fail to match the retention stability of kNN.

Self-Organizing Maps (SOMs)~\cite{som} present a compelling architecture that lies between these two extremes. As unsupervised models that project high-dimensional data onto a low-dimensional grid, SOMs rely on a competitive learning mechanism. Unlike fully connected networks where updates are global, SOMs update only the Best Matching Unit (BMU) and its immediate topological neighbors. This locality offers an intrinsic degree of resilience against catastrophic forgetting, as new information does not necessarily overwrite distal regions of the map.

However, this inherent resilience is insufficient for robust continual learning. As we demonstrate in this study, standard SOMs eventually succumb to forgetting when the data distribution shifts significantly; neurons previously specialized for old tasks are gradually pulled toward new data clusters, eroding prior representations. This limitation motivates the need for a mechanism that leverages the SOM's topological advantages while explicitly preventing the degradation of well-established knowledge.

\begin{figure*} \centering \includegraphics[width=\textwidth]{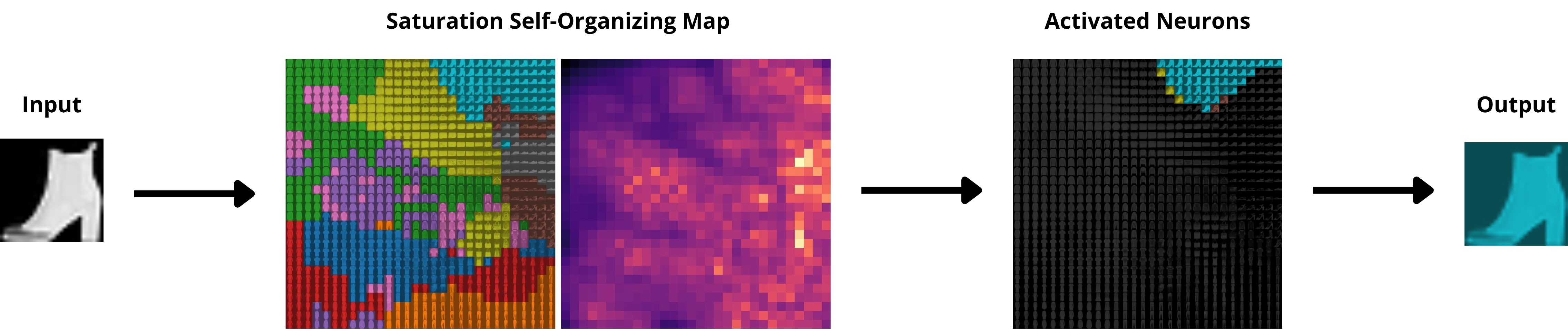} \caption{Symbolic visualization of SatSOM inference on the FashionMNIST dataset. An input image is processed to calculate activation levels for each neuron based on similarity. Label values associated with activated neurons are then aggregated to compute the final class probabilities.}\label{fig:inference} \end{figure*}

In this work, we propose the Saturation Self-Organizing Map (SatSOM), a novel extension of Self-Organizing Maps designed to enhance knowledge retention while directing learning signals toward underutilized model components. Building upon prior efforts to adapt SOMs for continual learning, such as the PROPRE framework~\cite{propre}, which demonstrated the potential of SOMs for knowledge retention, SatSOM introduces a saturation mechanism that tracks the “maturity” of individual neurons. This mechanism is based on a novel \textit{saturation} metric (conceptually distinct from traditional activation saturation in deep neural networks) that quantifies a neuron’s cumulative exposure to relevant training data. Once a neuron becomes saturated, its weight updates are restricted, effectively freezing critical learned patterns while allowing less-utilized regions of the map to remain plastic. By selectively preserving learned representations in this manner, SatSOM addresses the stability-plasticity dilemma central to catastrophic forgetting and achieves robust knowledge retention in dynamic environments without the infinite memory costs associated with kNN methods. A high-level overview of the inference process is provided in Figure~\ref{fig:inference}.

The primary contribution of this work is the demonstration that this saturation-based freezing mechanism significantly enhances knowledge retention. By explicitly preventing the overwriting of well-established representations, SatSOM achieves robustness superior to standard SOMs in dynamic environments. Furthermore, the principles underlying this mechanism provide a generalizable framework for stability. We posit that similar saturation-based constraints could be adapted for other neural architectures, potentially extending these benefits to more complex deep learning models.

We evaluate SatSOM through a rigorous experimental protocol involving two distinct sets of benchmarks. First, we provide a standard comparison against state-of-the-art approaches on image classification tasks. Second, we introduce a custom benchmark specifically designed for class-incremental learning across MNIST variants without data replay. This strict evaluation setting allows us to isolate model behavior and identify specific limitations in existing methods. Our findings suggest that the saturation mechanism offers a practical, interpretable, and effective solution for continual learning.

To facilitate reproducibility and further research, all model architectures and testing code used in this study are available in a public GitHub repository\footnote{https://github.com/Radinyn/satsom}.

\section{Related Work}

A comprehensive survey by Delange et al.~\cite{survey} details the landscape of continual learning, categorizing approaches into three primary families: regularization-based, rehearsal-based, and architectural or hybrid methods.

Regularization-based methods mitigate catastrophic forgetting by constraining weight updates to preserve knowledge from previous tasks. A representative example is Elastic Weight Consolidation (EWC)~\cite{EWC}, which penalizes changes to parameters deemed important for prior tasks. Conversely, rehearsal-based methods (or replay methods) retain a subset of past data to interleave with current training samples. For instance, iCaRL~\cite{icarl} combines a nearest-mean-of-exemplars classification rule with knowledge distillation to maintain feature representations in a class-incremental setting. Other efficient replay strategies include those proposed by Hayes et al.~\cite{memory_efficient_replay}.

The third category, architectural methods, mitigates forgetting by dynamically modifying the network structure or isolating task-specific parameters. Within this domain, Self-Organizing Maps (SOMs) have emerged as a potential tool for continual learning. Gepperth and Karaoguz~\cite{propre} demonstrated that SOMs can achieve competitive performance in continual learning scenarios, serving as a foundational inspiration for our approach. Further explorations into energy-based SOM training have also been conducted by Gepperth~\cite{energy_based}.

Several hybrid architectures have since integrated SOMs to enhance plasticity and stability. SOMPL, proposed by Bashivan et al.~\cite{SOMPL}, runs a SOM layer in parallel with a standard supervised network. The SOM clusters inputs into task contexts to generate an output mask, selectively routing inputs to relevant network modules without requiring explicit task labels. Similarly, DendSOM~\cite{DendSOM} draws inspiration from biological dendrites, employing a layer of multiple SOMs where each acts as a dendrite modeling a specific subregion of the input space.

Beyond the aforementioned categories, a complementary line of work focuses on detecting distributional shifts or concept drift in streaming data. A recent example is the approach proposed by Prowik et al.~\cite{drift_detection}, which detects drift based on measurements of changes in feature ranks. Such drift detection signals can subsequently be employed to adapt models online, similarly to Hoeffding Trees~\cite{Hoeffding}.

To address the challenges of online, task-free learning, Pourcel et al.~\cite{pourcel2022online} introduced Dynamic Sparse Distributed Memory (DSDM). This method utilizes a semi-distributed associative memory composed of partially overlapping clusters. DSDM dynamically evolves to model non-stationary data streams and employs a local density-based pruning technique to manage memory constraints effectively.

Most relevant to our work is the Continual SOM (CSOM) introduced by Vaidya et al.~\cite{CSOM}, which generalizes the SOM for online, task-free settings. CSOM embeds each neuron with a running variance and utilizes neuron-specific learning parameters. While SatSOM shares conceptual similarities with CSOM, it introduces several distinct features. First, unlike CSOM which maintains a running variance, SatSOM employs a standard Euclidean distance while achieving comparable performance. Second, SatSOM introduces a quantile threshold parameter to select only the most relevant neurons, thereby improving stability. Third, the streamlined architecture facilitates comprehensive ablation studies to identify key factors for future research, such as adapting the saturation concept to other models. Finally, we present a direct comparison with a kNN baseline, demonstrating that SatSOM outperforms both the baseline and CSOM on the selected benchmark.

\section{Methodology}\label{sec:architecture}


The \textit{Saturation Self-Organizing Map (SatSOM)} is founded on the principle that catastrophic forgetting can be mitigated by directing new learning toward underutilized components of the model, while preserving the stability of representations that are already significant for prediction. This approach is naturally facilitated by the topological grid structure of the SOM.

Implementation is achieved by assigning each neuron a specific learning rate and neighborhood radius. These parameters decay over time based on the cumulative magnitude of updates applied to the neuron, effectively stabilizing ("freezing") it once a specific threshold is reached. We define this state as \textit{saturation}—quantified here as the normalized difference between the initial and current learning rate, though it could explicitly be defined via the neighborhood radius as well.

\begin{figure}
\centering
\includegraphics[width=\columnwidth]{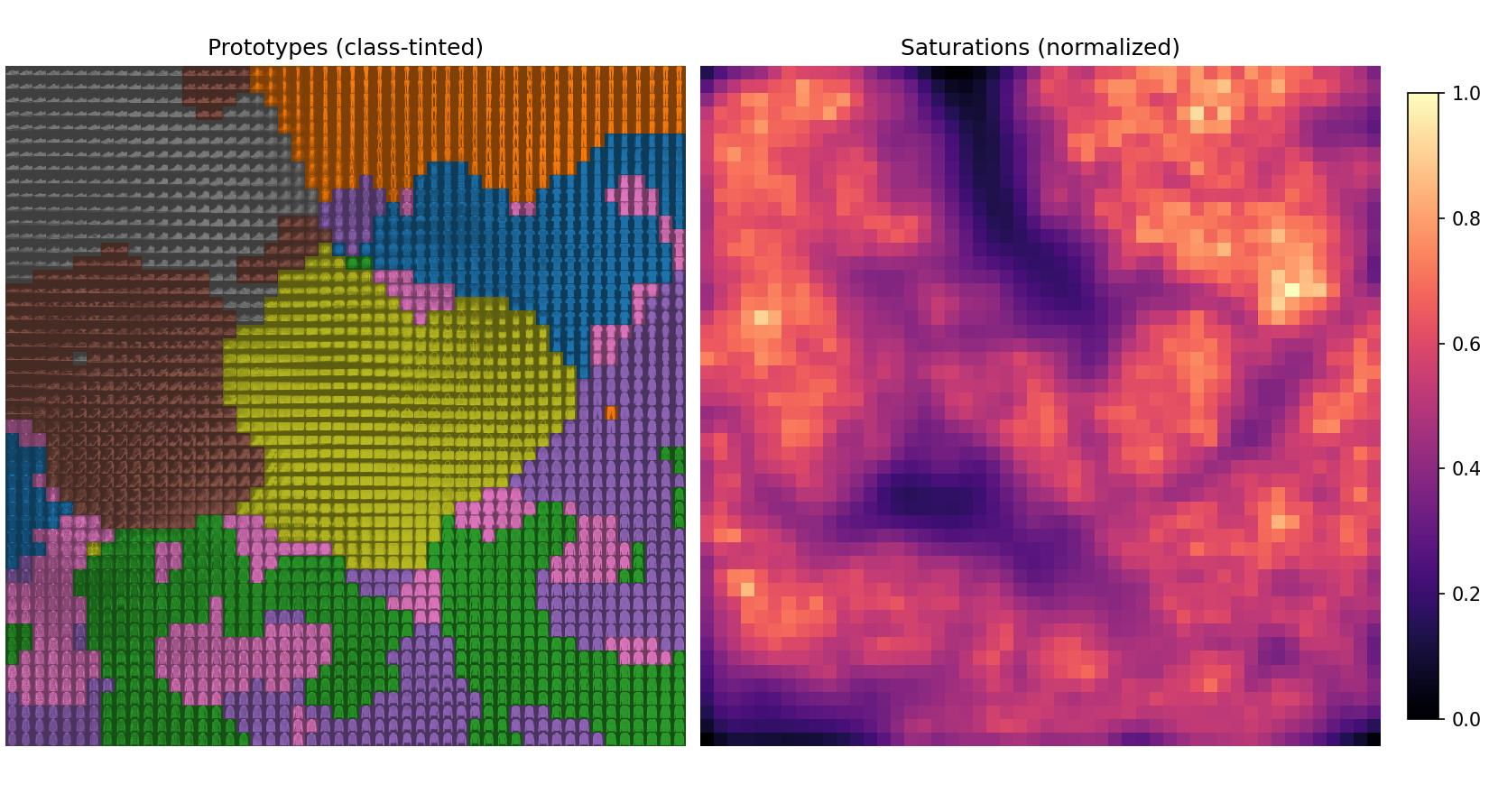}
\caption{SatSOM prototypes trained on FashionMNIST. Neuron weights are visualized as images and color-coded based on the class with the highest probability in their respective label vector.}\label{fig:example}
\end{figure}

SatSOM optimizes a set of prototypes (neuron weights) and associated label-prototypes. During inference, the model identifies the prototypes most similar to the input using a predefined distance metric (specifically, Euclidean distance) and aggregates their corresponding label-prototypes to generate a weighted output. This process is illustrated in Figure~\ref{fig:example}.

Our intuition is that the neighborhood radius decay guides new knowledge to the emptier parts of the map, while preserving the locality of changes. In contrast, learning rate decay works by preserving the more important neurons when no more space can be found. This view is further reinforced by the results of our ablation study (see Section~\ref{sec:ablation}).

\subsection{Definitions}\label{sec:definitions}

For simplicity, we assume that each SatSOM is a square map with a side length of $n$ neurons. We denote their total by $N=n^2$.

\begin{figure*}
\centering
\includegraphics[width=\textwidth]{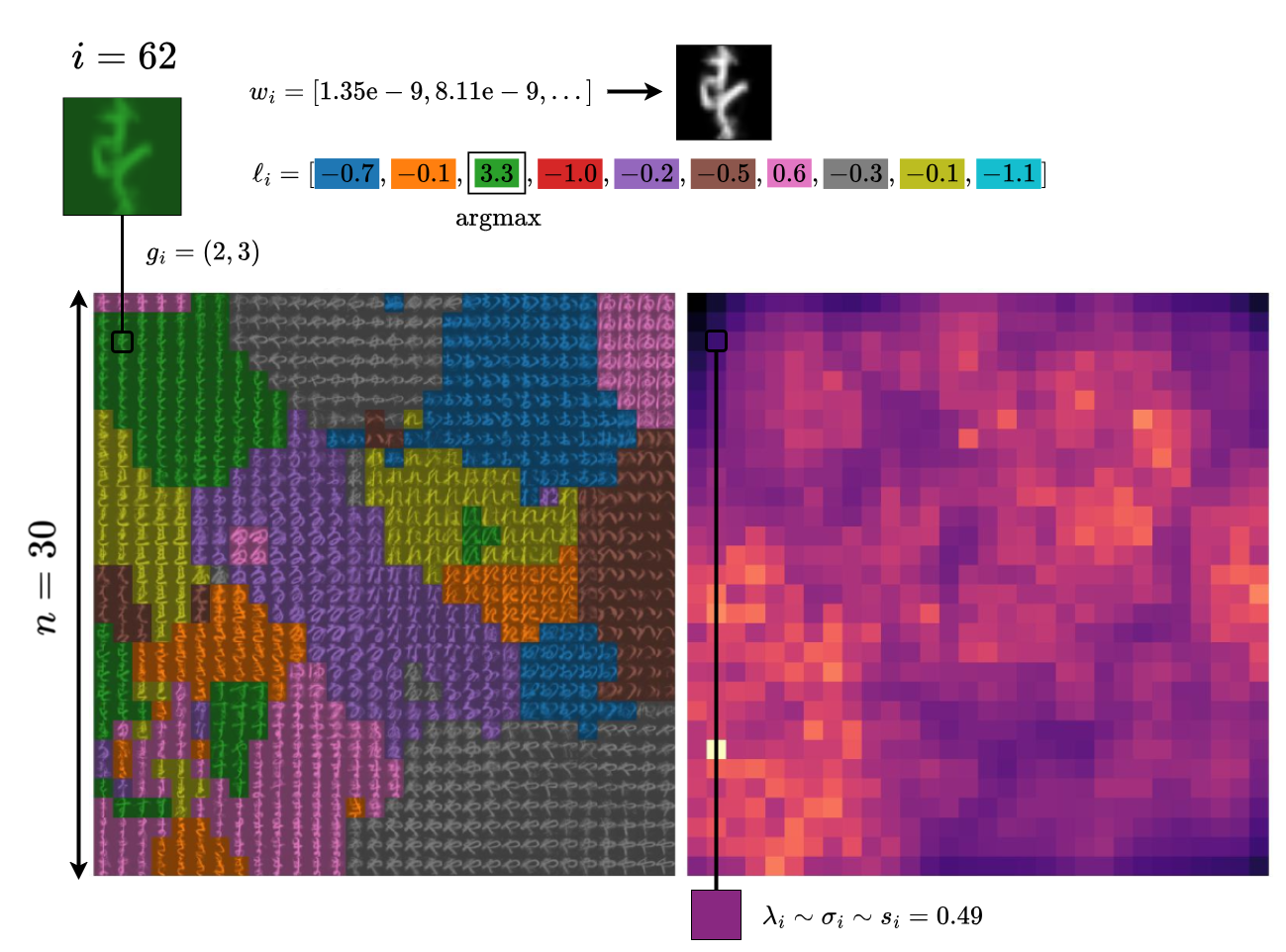}
\caption{Visualization of a SatSOM trained on the KMNIST dataset, annotated with key mathematical elements.}\label{fig:math}
\end{figure*}

Each neuron $i$ has its own prototype vector $w_i$ and prototype-label vector $\ell_i$. We denote its coordinates within the grid as $g_i$. To facilitate the saturation mechanism, two additional parameters are stored, the learning rate $\lambda_i$ and the neighborhood radius $\sigma_i$. We provide an intuitive visualization in Figure~\ref{fig:math}.

The model incorporates global hyperparameters $\lambda_0$ and $\sigma_0$ to initialize the learning rate and neighborhood radius of each neuron. These parameters decay exponentially during training at rates determined by $\alpha_\lambda$ and $\alpha_\sigma$ (Equation~\ref{eq:decay}), mirroring the decay schedule of traditional SOMs~\cite{som}.

For practical purposes, there are also two inference-time hyperparameters, $p$ and $q$. Our empirical evaluation across a range of configurations indicates that the model is largely insensitive to reasonable choices of these parameters. They are necessary for the model to work on unbalanced datasets, as described further.

Although SatSOM has many hyperparameters, in this work we show that specific combinations of those parameters are effective for most tasks.

Finally, we also define saturation of a neuron as:
\begin{equation}
s_i= \frac{\lambda_0 - \lambda_i}{\lambda_0},\quad i = 1, \dots, N.
\label{eq:saturation}
\end{equation}

This measure provides a compact indicator of how much the adaptive capacity of a neuron has diminished during training. By expressing the decay of the learning rate in normalized form, the saturation quantifies the extent to which a neuron has stabilized.

\subsection{Training}\label{sec:training}
The training procedure begins by presenting a single example $x$ together with its one‑hot-encoded label $y$. First, the Best‑Matching Unit (BMU) is found by computing Euclidean distances
\begin{equation}
d_i = \lVert x - w_i\rVert_2,
\end{equation}
and selecting its index:
\begin{equation}
b = \arg\min_i d_i.
\end{equation}
Around this BMU, each neuron $i$ acquires a neighborhood strength, based on the distance between its and BMU coordinates:
\begin{equation}
\theta_i = \exp\left(-\,\frac{\lVert g_i - g_b\rVert^2}{2\sigma_b\,\sigma_i}\right).
\end{equation}
Each prototype \(w_i\) is then nudged toward the sample by
\begin{equation}
w_i \leftarrow w_i + \lambda_i\,\theta_i\,(x - w_i).
\end{equation}
Concurrently, the label matrix is updated by first calculating the class probabilities (softmax):
\begin{equation}
p_{i,c} = \frac{\exp(\ell_{i,c})}{\sum_{c'} \exp(\ell_{i,c'})},
\end{equation}
where $c$ denotes class index,
computing the gradient \(\nabla \ell_i = p_i - \ell_i\), and applying
\begin{equation}
\ell_i \leftarrow \ell_i - \lambda_i\,\theta_i\,\nabla \ell_i.
\end{equation}
This is equivalent to using cross-entropy loss. Afterwards, both the per-neuron learning rate and radius decay multiplicatively according to the neighborhood strength:
\begin{equation}
\lambda_i \leftarrow \lambda_i \exp(-\alpha_\lambda\,\theta_i),  
\qquad  
\sigma_i \leftarrow \sigma_i \exp(-\alpha_\sigma\,\theta_i).
\label{eq:decay}
\end{equation}

\subsection{Inference}\label{sec:inference}
The model makes a prediction by creating a weighted average of select label-prototypes. A symbolic visualization is shown in Figure~\ref{fig:inference}.

We start with normalizing the distance ($\varepsilon>0$, a small constant added for numerical stability):
\begin{equation}
\tilde d_i = \frac{d_i - \min_j d_j}{\max_j d_j - \min_j d_j + \varepsilon}.
\end{equation}
Next, we disable under-saturated neurons, i.e. those that do not yet store any information:
\begin{equation}
\label{eq:epsilon}
\tilde d_i \leftarrow 1\quad\text{if }\quad s_i<\varepsilon.
\end{equation}
We compute a cut-off quantile threshold among the enabled neurons such that only the fraction $q$ of neurons closest to the input is used in the prediction:
\begin{equation}
\tau = \mathrm{quantile}\left(\{\tilde d_j : s_j \ge \varepsilon\},\, q\right),
\label{eq:threshold}
\end{equation}
and clamp any \(\tilde d_i\) exceeding \(\tau\) to unity:
\begin{equation}
\tilde d_i \leftarrow 1\quad \text{if } \tilde d_i > \tau.
\end{equation}
In effect, this procedure is analogous to selecting the nearest neighbors in a kNN classifier, as it prioritizes neurons in the local vicinity of the best-matching unit while mitigating the influence of more prevalent classes that constitute larger portions of the network, making the method robust on unbalanced datasets.

Finally, distances are converted into neuron \textit{proximities} via the exponent ($p$), sharpening the fall-off:
\begin{equation}
h_i = \left(1 - \tilde d_i\right)^{p}.
\end{equation}
We use the label matrix to compute the final prediction, weighted by proximity:
\begin{equation}
\hat y = \frac{1}{N} \sum_{i=1}^N h_i \ell_i.
\end{equation}

\section{Evaluation}

The primary objective of our experimental analysis is to rigorously assess the knowledge retention capabilities of SatSOM in strictly online, class-incremental settings. We adopt a two-tiered evaluation strategy: first, we benchmark the model on raw pixel data using standard datasets (FashionMNIST, KMNIST) to validate its fundamental learning dynamics. Second, we extend the evaluation to complex, high-dimensional feature spaces (CIFAR-10, CIFAR-100, CORe50) using pre-trained embeddings, isolating the model's ability to manage stability-plasticity trade-offs when feature extraction is handled externally.

\subsection{Metrics}

We employ standard continual learning metrics to quantify performance.
Given a prediction function \( f : \mathcal{X} \to \mathcal{Y} \) and a dataset 
\( \mathcal{D} = \{(x_i, y_i)\}_{i=1}^{N} \), the average accuracy is:
\begin{equation}
\label{eq:accuracy}
\mathrm{ACC}(f, \mathcal{D})
    = \frac{1}{N}\sum_{i=1}^{N}
        \mathbf{1}\!\left( f(x_i) = y_i \right),
\end{equation}
where \( \mathbf{1}(\cdot) \) is the indicator function.

We report the Average Accuracy (ACC) over all classes at the end of the entire training sequence.

Let the scenario consist of tasks 
\( \{T_1,\dots,T_K\} \).  
We denote by \( f_k \) the model after training on task \( T_k \).

To measure the stability of the most recently acquired knowledge, we report Last Accuracy (LA), defined as the performance on the final task immediately after learning it
\begin{equation}
\mathrm{LA} = \mathrm{ACC}(f_K, T_K).
\end{equation}

We quantify forgetting using the Forgetting Measure which quantifies the average degradation in performance on previously learned tasks after completing all tasks.  
Let
\(
A_{k}^{\max} = \max_{j \in \{k,\dots,K\}} \mathrm{ACC}(f_j, T_k)
\)
be the best accuracy achieved on task \( T_k \) at any point during training.  
Then the forgetting measure is
\begin{equation}
\mathrm{FM}
    = \frac{1}{K-1} \sum_{k=1}^{K-1}
        \left( A_{k}^{\max} - \mathrm{ACC}(f_K, T_k) \right).
\end{equation}
Positive values indicate forgetting; zero indicates no forgetting.

Additionally, Backward Transfer measures the influence of new learning on old tasks.  
Using the same notation as above, BWT is defined as
\begin{equation}
\label{eq:bwt}
\mathrm{BWT}
    = \frac{1}{K-1} \sum_{k=1}^{K-1}
        \left( \mathrm{ACC}(f_K, T_k) - \mathrm{ACC}(f_k, T_k) \right).
\end{equation}
Positive values indicate improvement of old tasks and negative values indicate interference.

\subsection{Direct Evaluation}\label{sec:testing}

We first evaluate SatSOM in a challenging class-incremental setting where the model must learn directly from raw pixel intensities. We utilize two 10-class benchmarks: FashionMNIST~\cite{FashionMNIST} and KuzushijiMNIST (KMNIST)~\cite{KMNIST}, using a 70-30 train-validation split. Training is organized into distinct phases, where each phase presents samples from a single class exclusively. Once a phase concludes, that class is never revisited. This protocol—single-pass, online, class-incremental learning—represents a severe test of catastrophic forgetting.

We benchmark against two baselines: a deterministic kNN ($k=5$), which serves as an upper bound for retention (due to its full memory of the dataset), and the DSDM architecture~\cite{pourcel2022online}. Notably, all models are evaluated in a strict single-pass setting. The hyperparameters are detailed in Table~\ref{tab:hyperparams}.

\begin{table}
\centering
\begin{tabular}{@{}llll@{}}
\toprule
\textbf{kNN} & \textbf{SatSOM} & \textbf{DSDM} & \textbf{Value} \\
\midrule
$k$ & -- & -- & 5 \\
-- & $n$ & -- & 100 \\
-- & $\lambda_0$ & -- & 0.5 \\
-- & $\sigma_0$ & -- & 10 \\
-- & $\alpha_\lambda$ & -- & 0.01 \\
-- & $\alpha_\sigma$ & -- & 0.2 \\
-- & $p$ & -- & 10 \\
-- & $q$ & -- & 0.001 \\
-- & -- &  $\beta$ & 2 \\
-- & -- &  $\lambda$ & 0.01 \\
\bottomrule
\end{tabular}
\caption{Hyperparameters used in the experimental evaluation.}\label{tab:hyperparams}
\end{table}

\begin{figure*}[!htb]
\centering
\includegraphics[width=\textwidth]{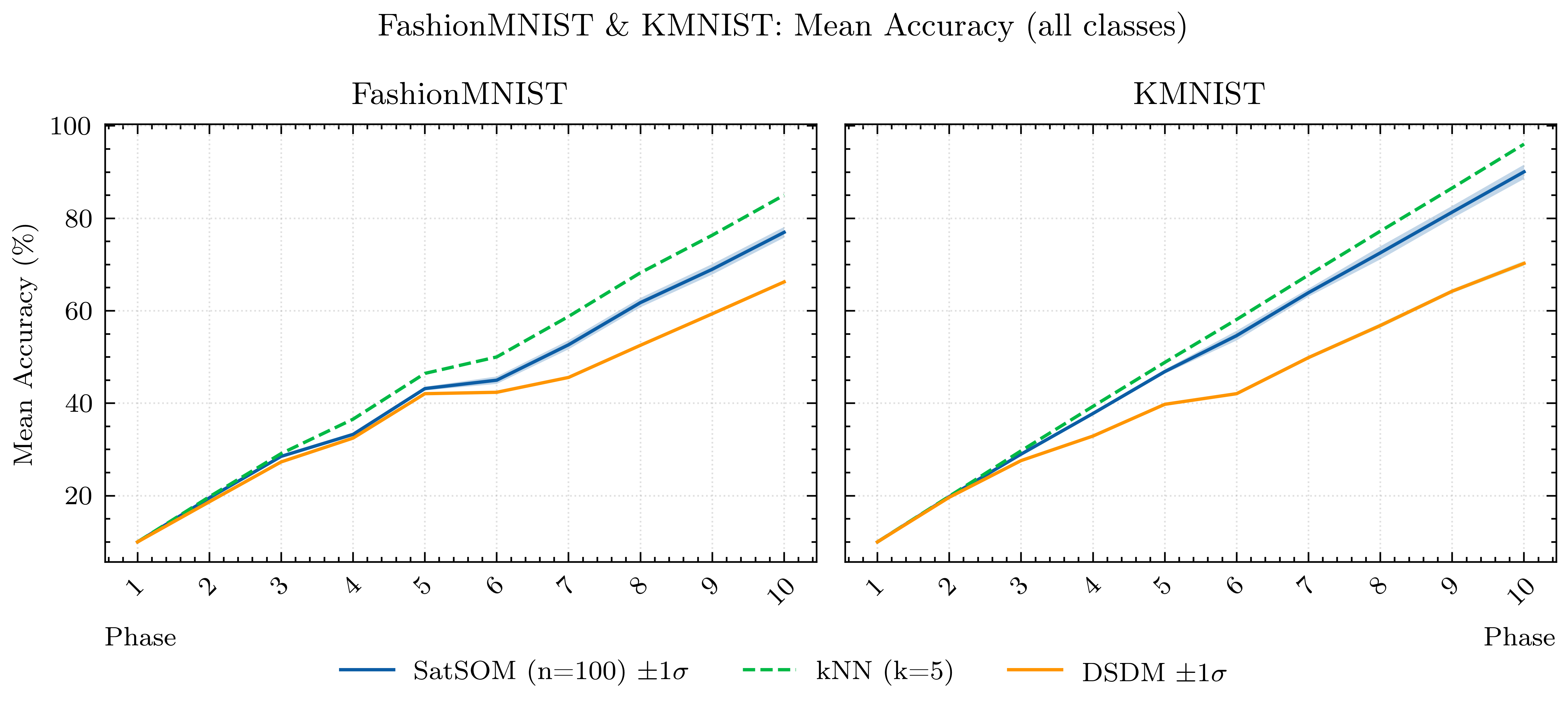}
\caption{Evolution of mean accuracy across all observed classes throughout the 10 training phases.}\label{fig:mnist_mean_all}
\end{figure*}

\begin{figure*}[!htb]
\centering
\includegraphics[width=0.8\textwidth]{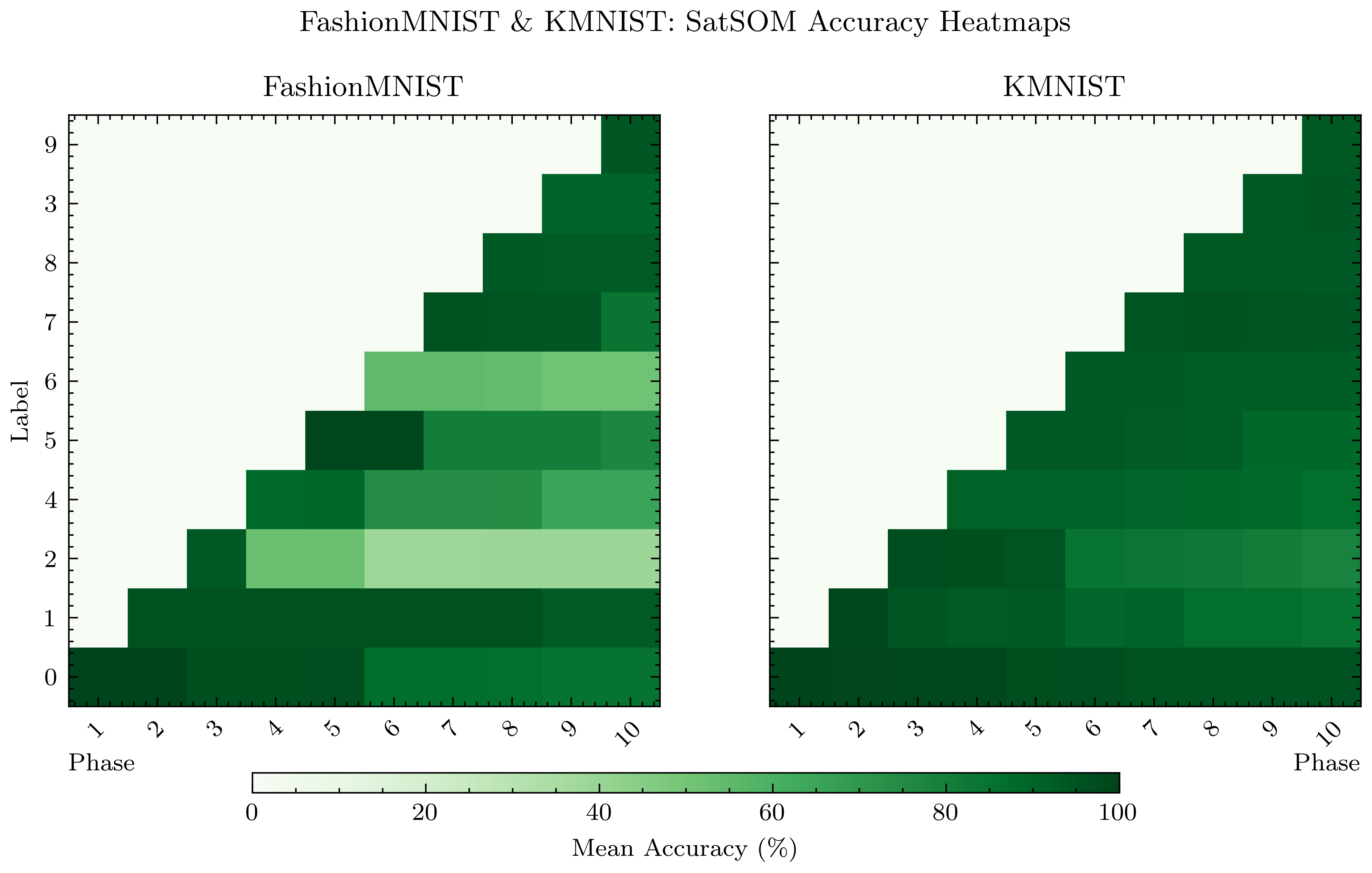}
\caption{Heatmap of mean SatSOM accuracy per class. Performance fluctuations are likely attributable to inter-class visual similarities typical in these datasets.}\label{fig:mnist_heat}
\end{figure*}

The performance trajectory across all 10 phases is illustrated in Figure~\ref{fig:mnist_mean_all}, with class-specific breakdowns in Figure~\ref{fig:mnist_heat}. Quantitative comparisons are summarized in Table~\ref{tab:soms}.

SatSOM demonstrates robust long-term retention, exhibiting a stability profile closer to the kNN baseline than to the plastic DSDM architecture. It outperforms DSDM significantly on both datasets. Crucially, SatSOM achieves this without requiring task identifiers or boundaries, highlighting its versatility for autonomous scenarios.

\begin{figure*}
\includegraphics[width=\textwidth]{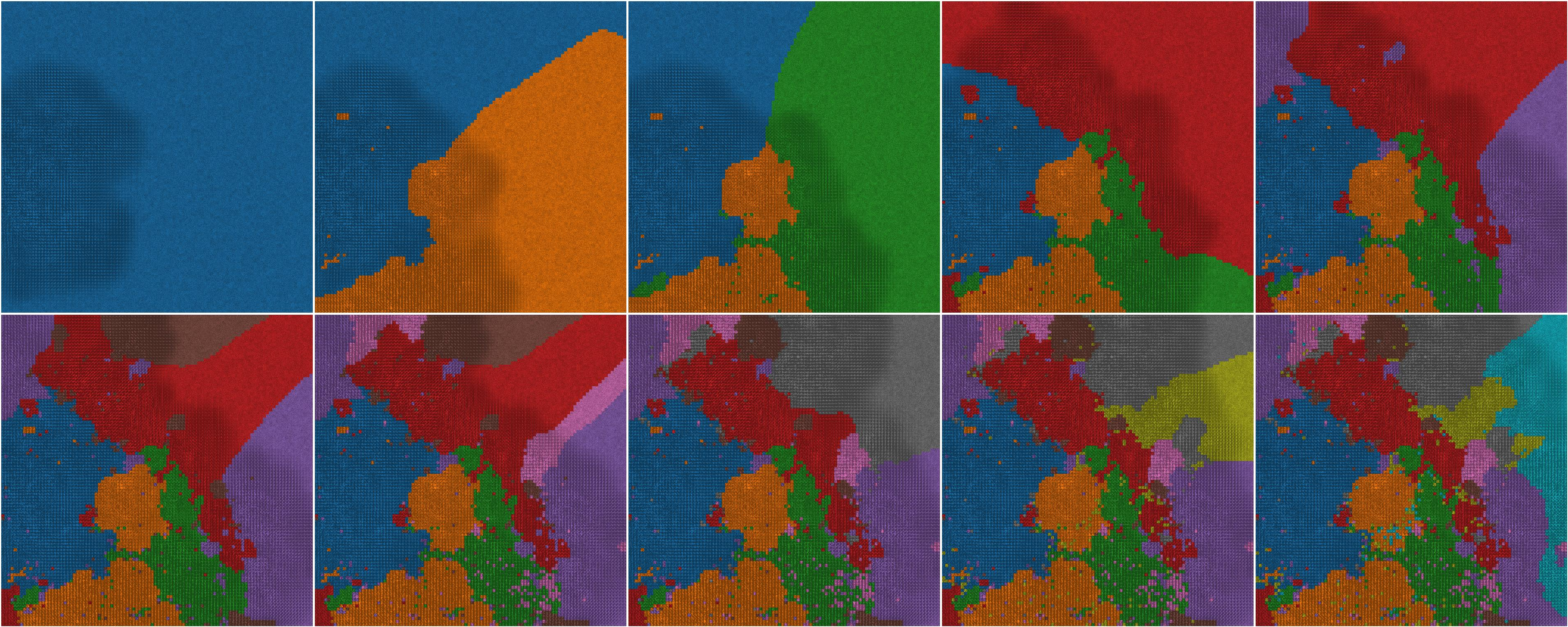}
\caption{Evolution of prototypes during a single training pass on the KMNIST dataset (ordered left to right, top to bottom).}\label{fig:prototypes}
\end{figure*}

\begin{figure*}
\centering
\includegraphics[width=\textwidth]{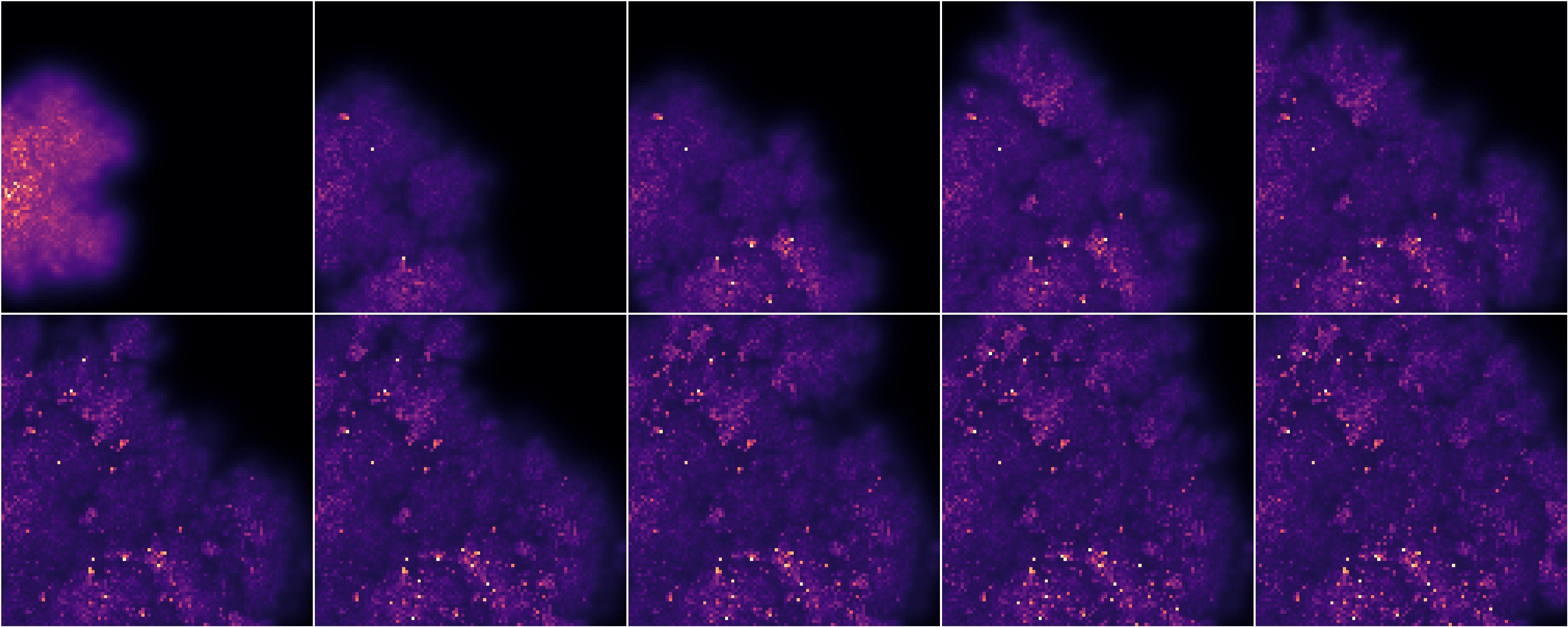}
\caption{Evolution of neuron saturation corresponding to the run in Figure~\ref{fig:prototypes}..}\label{fig:saturations}
\end{figure*}

To visualize the internal dynamics, Figures~\ref{fig:prototypes} and \ref{fig:saturations} display the evolution of prototypes and neuron saturation during a KMNIST run. The relatively low saturation levels suggest that the network retains sufficient plasticity to accommodate future classes if needed, a balance controlled by the choice of hyperparameters.




Table~\ref{tab:soms} compares SatSOM against other SOM-based methods. SatSOM achieves the highest accuracy among alike approaches. While DendSOM records favorable FM and BWT scores, this is an artifact of its low baseline accuracy (FM and BWT are relative metrics; a model that learns little has little to forget). SatSOM, conversely, maintains high absolute accuracy with minimal forgetting.

\begin{table*}[!htb]
\centering
\begin{tabular}{l l c c c c}
\toprule
Dataset & Method & ACC (↑) & LA (↑) & FM (↓) & BWT (↑) \\
\midrule

\multirow{6}{*}{FashionMNIST (step 1)}
& kNN         & 85.03 & 91.77 & 7.49 & -7.49 \\
& DSDM & $66.25 \pm 0.28$ & $80.37 \pm 0.36$ & $15.70 \pm 0.32$ & $15.69 \pm 0.33$ \\
\cmidrule(lr){2-6}
& vanilla SOM & $32.07 \pm 0.03$ & $57.52 \pm 0.03$ & $37.01 \pm 0.03$ & $-28.27 \pm 0.02$ \\
& DendSOM     & $16.62 \pm 0.03$ & $27.36 \pm 0.03$ & \textbf{11.34 $\pm$ 0.03} & \textbf{-11.92 $\pm$ 0.03} \\
& CSOM        & $75.13 \pm 2.68$ & $88.23 \pm 0.56$ & $13.11 \pm 2.93$ & $-14.56 \pm 3.25$ \\
& \textbf{SatSOM}      & \textbf{76.98 $\pm$ 1.15} & \textbf{90.30 $\pm$ 0.54} & $14.93 \pm 1.67$ & $-14.79 \pm 1.66$ \\
\midrule

\multirow{6}{*}{KMNIST (step 1)}
& kNN         & 96.05 & 97.65 & 1.79 & -1.79 \\
& DSDM & $70.26 \pm 0.56$ & $73.49 \pm 0.61$ & $3.58 \pm 0.37$ & $-3.58 \pm 0.37$ \\
\cmidrule(lr){2-6}
& vanilla SOM & $23.65 \pm 0.03$ & $54.00 \pm 0.05$ & $35.28 \pm 0.01$ & $-33.79 \pm 0.02$ \\
& DendSOM     & $10.95 \pm 0.01$ & $12.92 \pm 0.01$ & \textbf{2.02 $\pm$ 0.00} & \textbf{-2.18 $\pm$ 0.01} \\
& CSOM        & $81.60 \pm 2.82$ & $88.61 \pm 1.60$ & $7.04 \pm 1.54$ & $-7.78 \pm 1.73$ \\
& \textbf{SatSOM}      & \textbf{90.06 $\pm$ 1.55} & \textbf{94.74 $\pm$ 0.62} & $5.42 \pm 1.64$ & $-5.20 \pm 1.67$ \\
\bottomrule
\end{tabular}
\caption{Comparison of SatSOM and other selected continual learning methods. Reference values for vanilla SOM, DendSOM, and CSOM are sourced from~\cite{CSOM}.}\label{tab:soms}
\end{table*}

\subsection{Evaluation with Pre-trained Backbone}

To rigorously test the scalability of SatSOM, we extend our evaluation to Latent Rehearsal settings. Following protocols similar to DSDM~\cite{pourcel2022online}, we freeze a ResNet50 backbone pre-trained on ImageNet and perform continual learning solely on the generated embeddings. While this simplifies the feature extraction burden, it isolates the plasticity-stability trade-off of the classification mechanism itself. It should be noted that this setting inherently favors methods relying on static similarity (like kNN) because the feature space is fixed.

For these experiments, we utilize the hyperparameters from Table~\ref{tab:hyperparams}, with the map size $n$ increased to $250$ to accommodate the higher complexity of the datasets.

We report results on CIFAR-10 and CIFAR-100 in class-incremental settings with varying step sizes (Table~\ref{tab:cifar}). SatSOM demonstrates competitive performance, marginally outperforming DSDM on CIFAR-10, though it trails on the more diverse CIFAR-100. Additionally, on the CORe50 benchmark~\cite{core50} (Table~\ref{tab:core}), tested in the Multi-Task New Classes (MT-NC) setting, SatSOM achieves respectable accuracy, although it does not yet match state-of-the-art specialized architectures.

\begin{table*}[t]
\centering
\begin{tabular}{l l c c c c}
\toprule
Dataset & Method & ACC (↑) & LA (↑) & FM (↓) & BWT (↑) \\
\midrule
\multirow{3}{*}{CIFAR-10 (step 1)}
& kNN & \textbf{86.38} & \textbf{92.81} & \underline{$7.15$} & \underline{$-7.15$} \\
& DSDM & $78.30 \pm 0.42$ & $84.13 \pm 0.36$ & \textbf{6.48 $\pm$ 0.25} & \textbf{-6.48 $\pm$ 0.25} \\
& \textbf{SatSOM} & \underline{$79.83 \pm 0.48$} & \underline{$92.57 \pm 0.52$} & $14.80 \pm 0.65$ & $-14.15 \pm 0.67$ \\
\midrule
\multirow{3}{*}{CIFAR-100 (step 2)} 
& kNN & \textbf{61.29} & \textbf{72.58} & \underline{11.52} & \underline{-11.52} \\
& DSDM & \underline{$51.91 \pm 0.36$} & $61.55 \pm 0.33$ & \textbf{9.86 $\pm$ 0.34} & \textbf{-9.84 $\pm$ 0.35} \\
& \textbf{SatSOM} & $44.34 \pm 0.46$ & \underline{$61.92 \pm 1.67$} & $20.44 \pm 1.50$ & $-17.94 \pm 1.55$ \\
\midrule
\multirow{3}{*}{CIFAR-100 (step 5)} 
& kNN & \textbf{61.29} & \textbf{71.95} & \underline{11.22} & \underline{-11.22} \\
& DSDM & \underline{$56.19 \pm 0.58$} & \underline{$63.75 \pm 0.77$} & \textbf{7.96 $\pm$ 0.27} & \textbf{-7.96 $\pm$ 0.27} \\
& \textbf{SatSOM} & $44.94 \pm 1.13$ & $63.03 \pm 1.40$ & $19.63 \pm 1.97$ & $-19.05 \pm 2.02$ \\
\bottomrule
\end{tabular}
\caption{Results on CIFAR-10 and CIFAR-100 in a class-incremental setting.}\label{tab:cifar}
\end{table*}

\begin{table}[t]
\centering
\begin{tabular}{l l c}
\toprule
Dataset & Method & ACC (↑) \\
\midrule
\multirow{3}{*}{CORe50}
& kNN   & \textbf{72.48} \\
& DSDM  & \underline{$71.48 \pm 0.26$} \\
& \textbf{SatSOM}   & $62.79 \pm 2.22$ \\
\bottomrule
\end{tabular}
\caption{Results on the CORe50 benchmark.}\label{tab:core}
\end{table}


\section{Ablation Study}\label{sec:ablation}

To isolate the impact of key algorithmic components, we conducted an ablation study using the FashionMNIST dataset. We replicated the experimental protocol from Section~\ref{sec:testing}, while selectively disabling specific hyperparameters. All other settings were maintained as specified in Table~\ref{tab:hyperparams}, and each configuration was evaluated across 10 independent runs.

\subsection{Learning Rate Decay}
Inhibition of $\alpha_\lambda$ by setting it to zero required additional change to avoid the quantile threshold disabling all the neurons (see Equation~\ref{eq:threshold}). For this test we temporarily re-defined saturation (see Equation~\ref{eq:saturation}) on the basis of $\sigma$ instead of $\lambda$:

\begin{equation}
s_i= \frac{\sigma_0 - \sigma_i}{\sigma_0},\quad i = 1, \dots, N.
\end{equation}

\begin{figure}[!h]
\centering
\includegraphics[width=0.9\columnwidth]{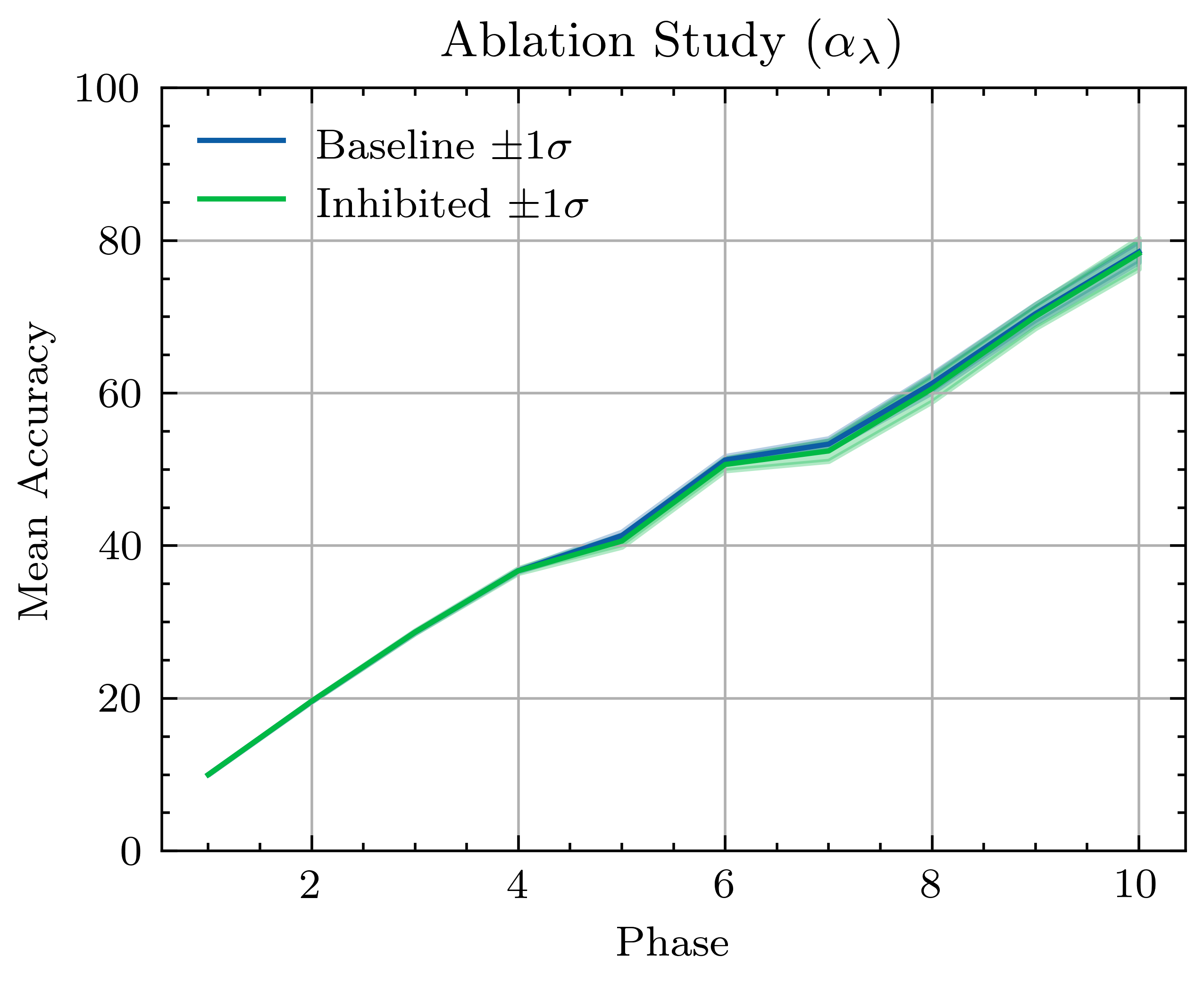}
\caption{Mean SatSOM accuracy (FashionMNIST) comparison between the base model and the inhibited model with $\alpha_\lambda=0$. The plot shows no impact due to the model's effectively unconstrained memory ($n=100$).}\label{fig:ablation_lr}
\end{figure}

\begin{figure}[!h]
\centering
\includegraphics[width=0.9\columnwidth]{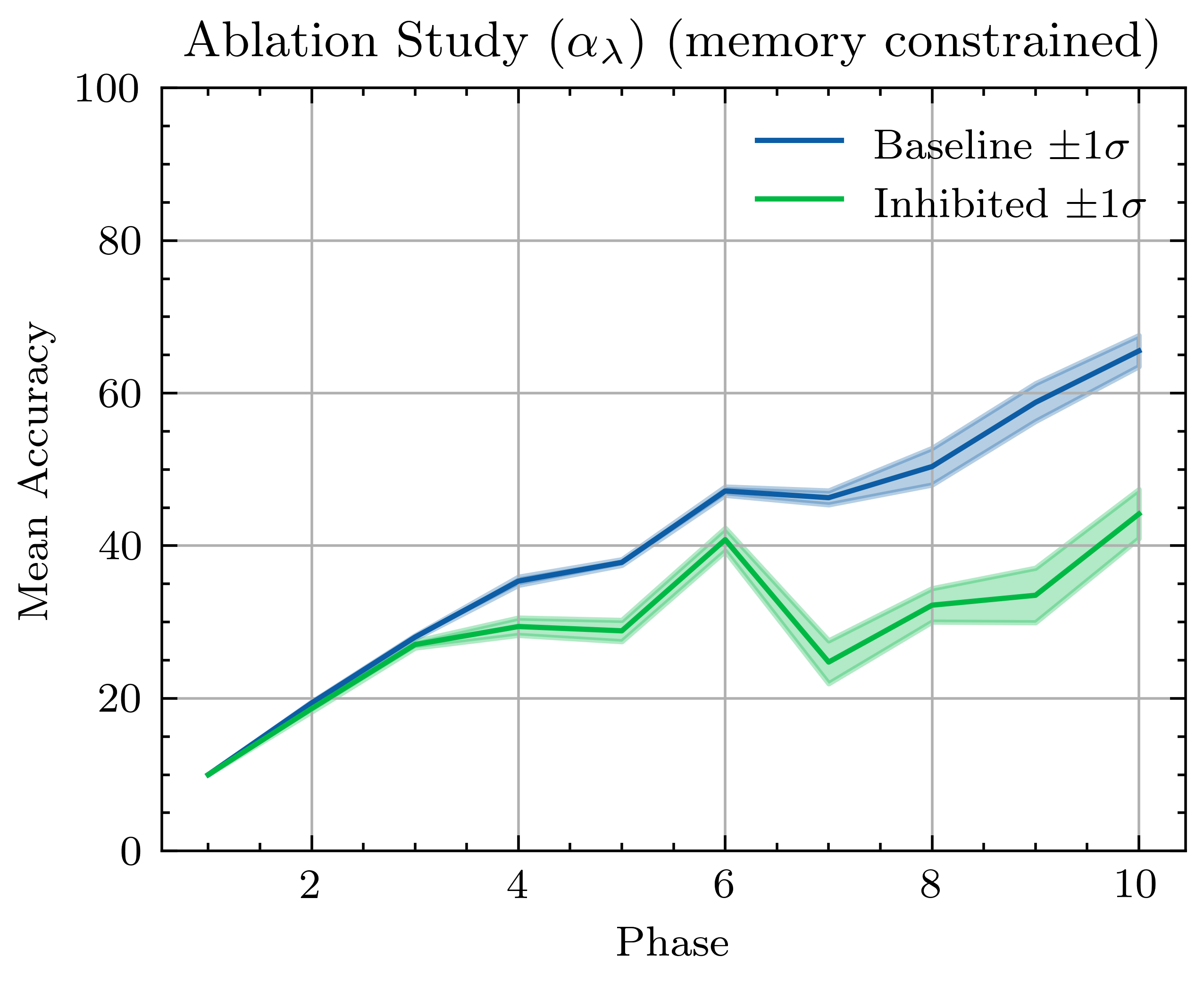}
\caption{Mean SatSOM accuracy (FashionMNIST) comparison between the base model and the inhibited model with $\alpha_\lambda=0$ in a memory constrained environment ($n=20$). The effect of $\alpha_\lambda$ is clearly visible.}\label{fig:ablation_lr_small}
\end{figure}

We also ran this test for two map sizes, the common $n=100$ (Figure~\ref{fig:ablation_lr}) and memory constrained $n=20$ (Figure~\ref{fig:ablation_lr_small}). The purpose of this is to show the situation in which learning rate decay comes into play.

The effect of $\alpha_\lambda$ is noticeable only in the $n=20$ example. This behavior showcases $\alpha_\lambda$ as an important hyperparameter preventing knowledge loss in memory constrained environments and reinforces $\alpha_\sigma$ as the main factor when SatSOM is not yet over-learned.

\subsection{Neighborhood Radius Decay}

\begin{figure}[thbp]
\centering
\includegraphics[width=\columnwidth]{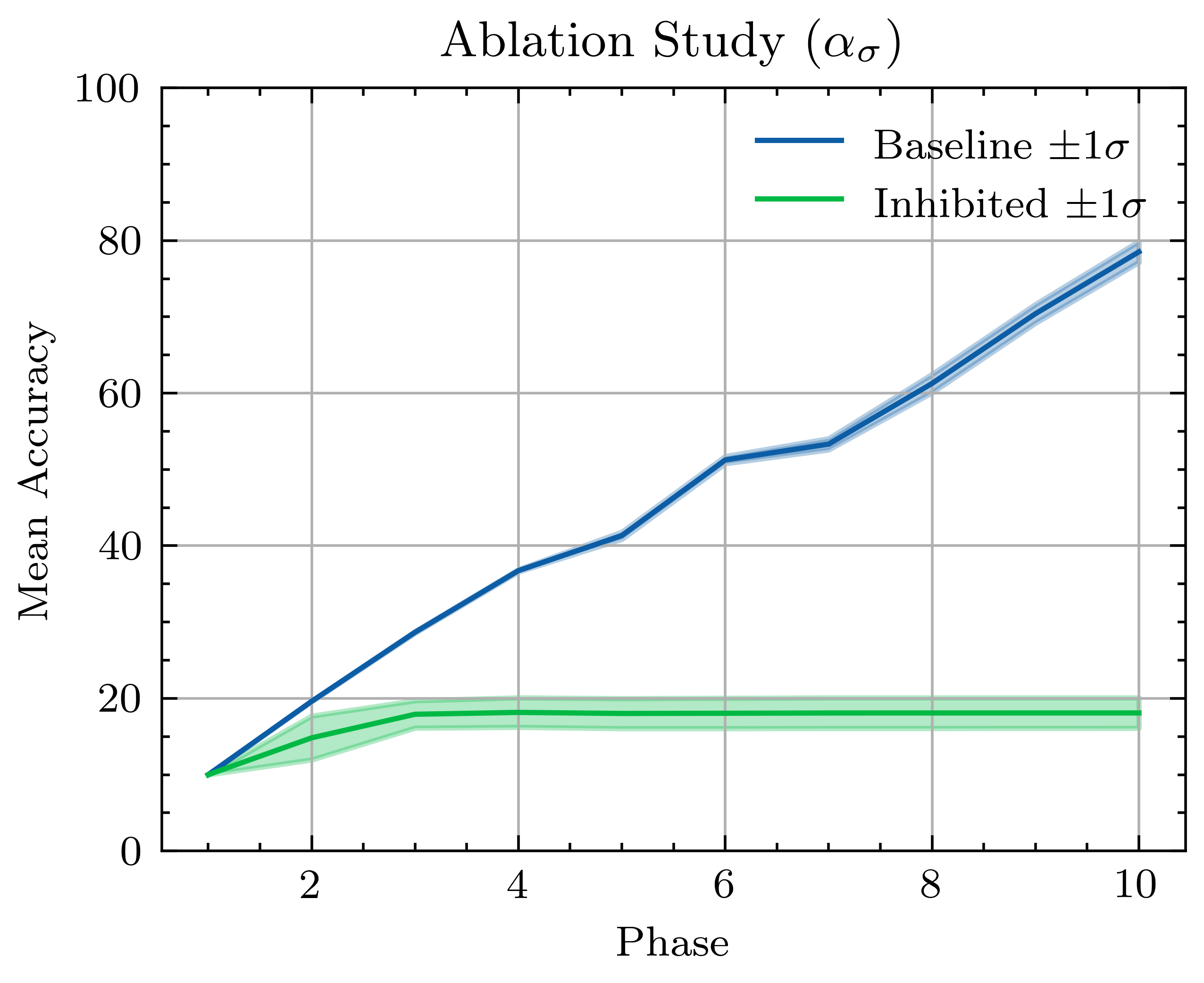}
\caption{Mean SatSOM accuracy (FashionMNIST) comparison between the base model and the inhibited model with $\alpha_\sigma=0$. There is a visible cut-off of accuracy in the inhibited model.}\label{fig:ablation_lr_sigma}
\end{figure}

\begin{figure}[thbp]
\centering
\includegraphics[width=\columnwidth]{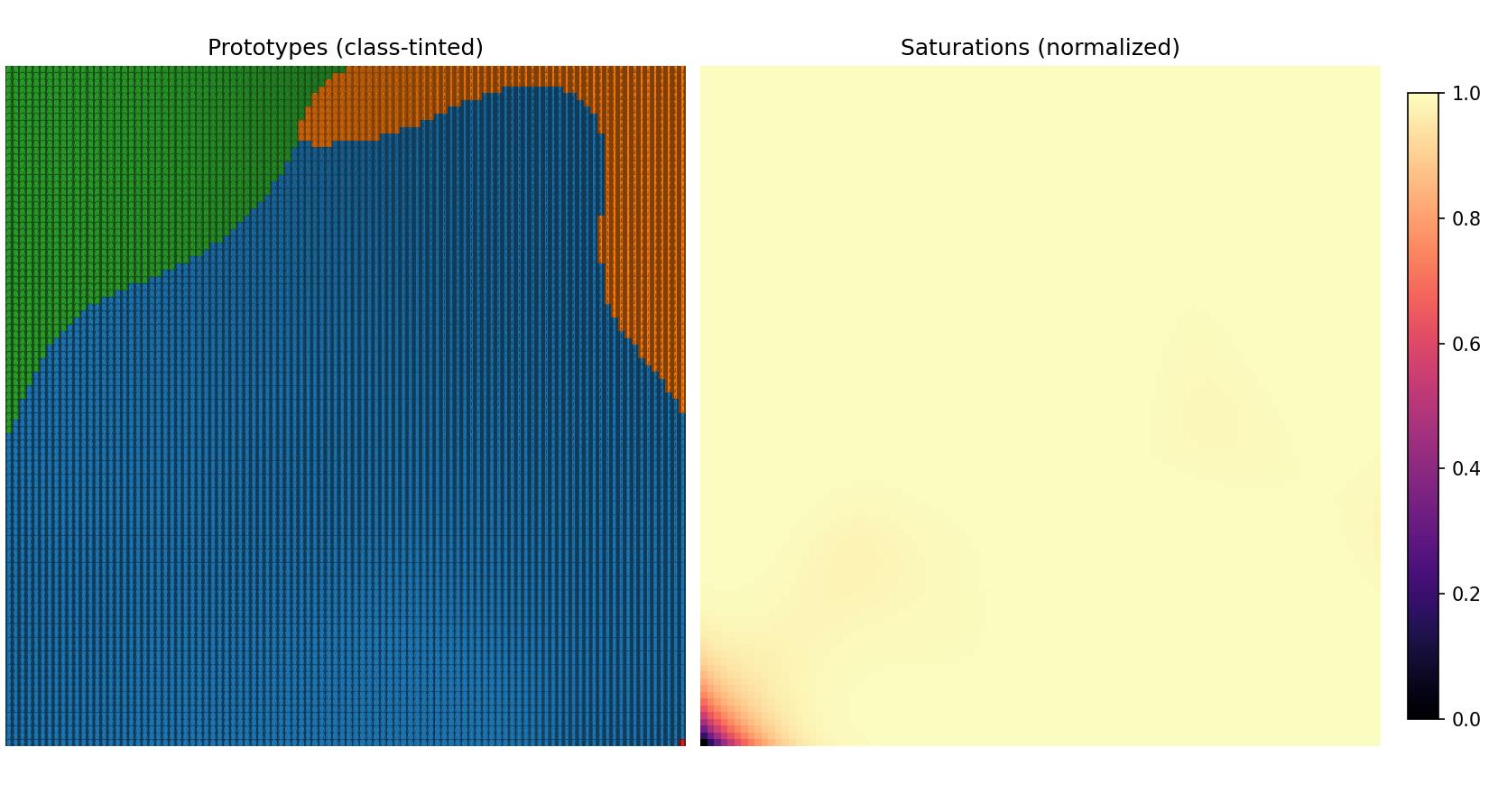}
\caption{Example SatSOM after 10 phases with inhibited neighborhood radius decay. Only 3 classes are present.}\label{fig:ablation_lr_sigma_map}
\end{figure}

In the next test, we set $\alpha_\sigma$ to $0$. Although the model demonstrates some knowledge retention, it quickly loses its ability to learn new classes (Figure~\ref{fig:ablation_lr_sigma}). The ones on which the model was already trained already occupy too much of the map (see Figure~\ref{fig:ablation_lr_sigma_map}). This also shows that SatSOM's memory is strictly limited by the choice of n.

\subsection{Quantile Threshold}
\begin{figure}[thbp]
\centering
\includegraphics[width=\columnwidth]{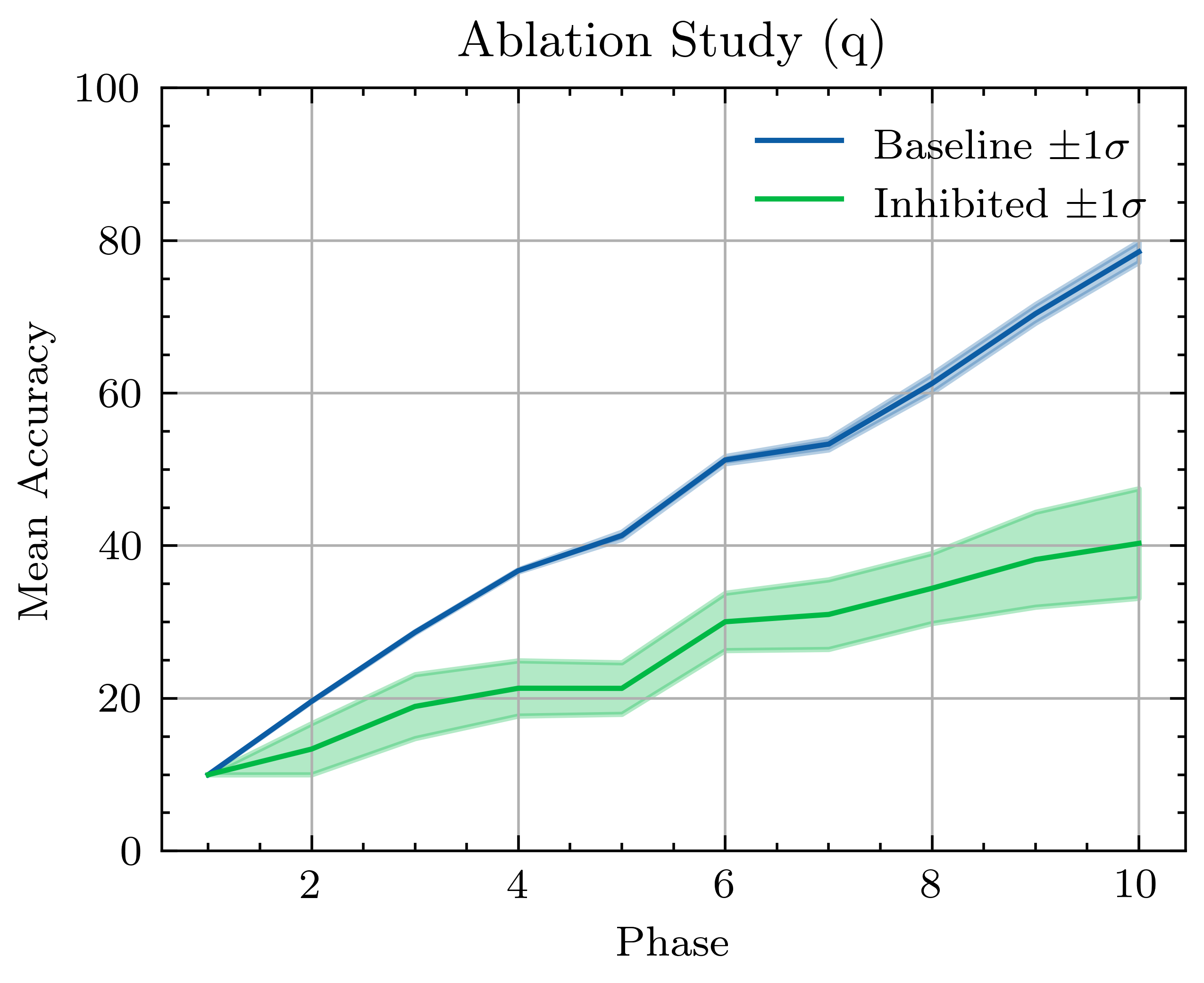}
\caption{Mean SatSOM accuracy (FashionMNIST) comparison between the base model and the inhibited model with $q=1$. The base model achieved superior results in all the runs.}\label{fig:ablation_quantile}
\end{figure}

Inhibiting q by setting it to 1 diminishes the model performance by including many irrelevant neurons in the final label calculation (Figure~\ref{fig:ablation_quantile}). SatSOM output becomes more sensitive to class sizes, as their relative size on the map weighs on the output, increasing the standard deviation.

\section{Hyperparameter Analysis}

We conducted a basic hyperparameter grid search during the initial prototyping phase to gain additional insights and intuitions. The exact grid of tested parameter values is presented in Table~\ref{tab:satsom_grid}.

\begin{table}
\centering
\label{tab:satsom_grid}
\begin{tabular}{@{}ll@{}}
\toprule
\textbf{Parameter} & \textbf{Values} \\
\midrule
$n$ & 50, 100 \\
$\lambda_0$ & 0.5 \\
$\sigma_0$ & $\frac{n}{2}$ \\
$\alpha_\lambda$ & 0.01 \\
$\alpha_\sigma$ & $10^{-3}$, $10^{-2}$, $10^{-1}$, 1 \\
$p$ & 10 \\
$q$ & \textbf{$10^{-5}$, $10^{-4}$, $10^{-3}$, $10^{-2}$} \\
\bottomrule
\end{tabular}
\caption{Hyperparameter search value grid.}
\end{table}

\begin{figure}
\centering
\includegraphics[width=\columnwidth]{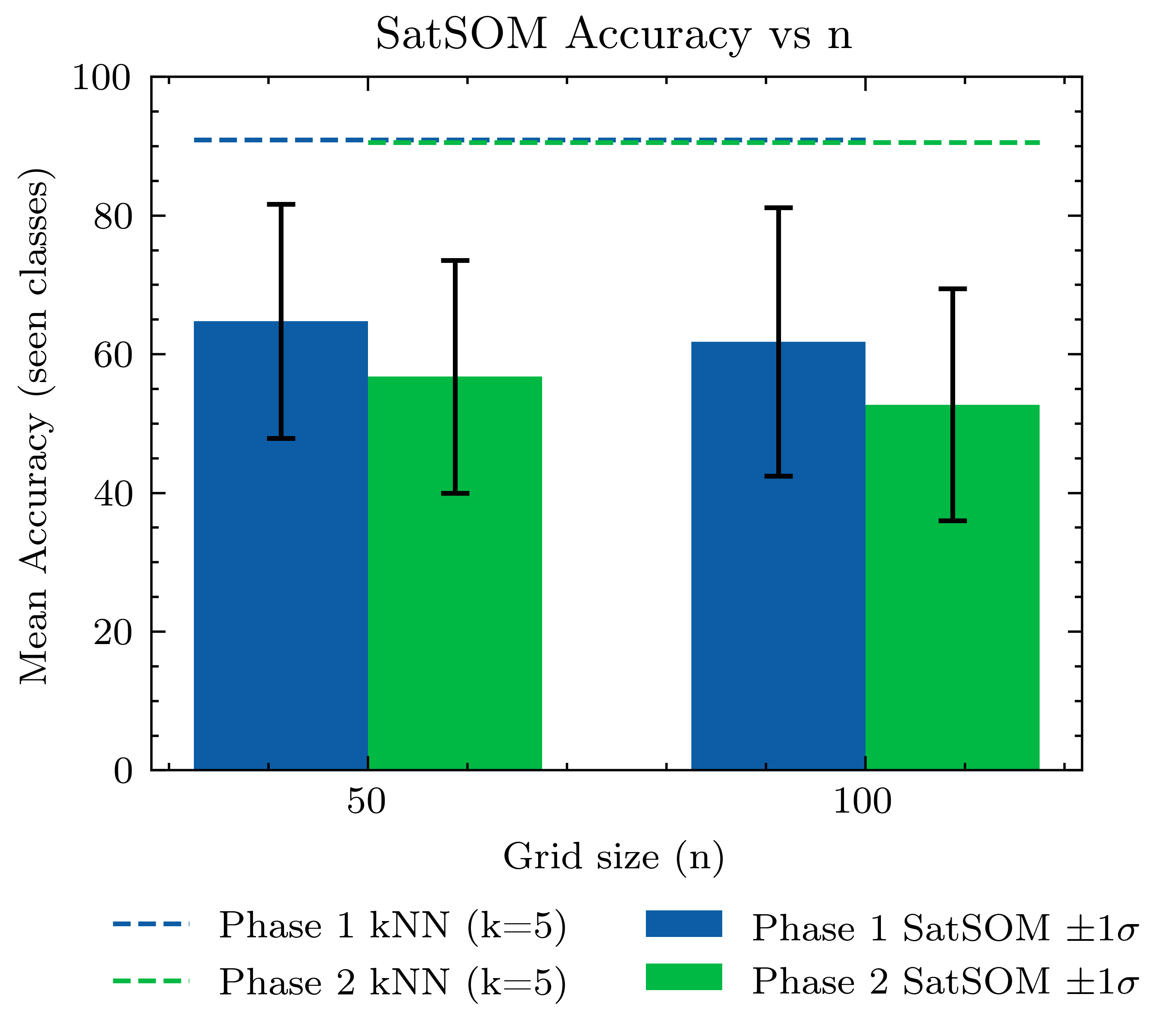}
\caption{Mean SatSOM accuracy with respect to the $n$ hyperparameter. There is no clear difference between the two values.}\label{fig:hyper_n}
\end{figure}

\begin{figure}
\centering
\includegraphics[width=\columnwidth]{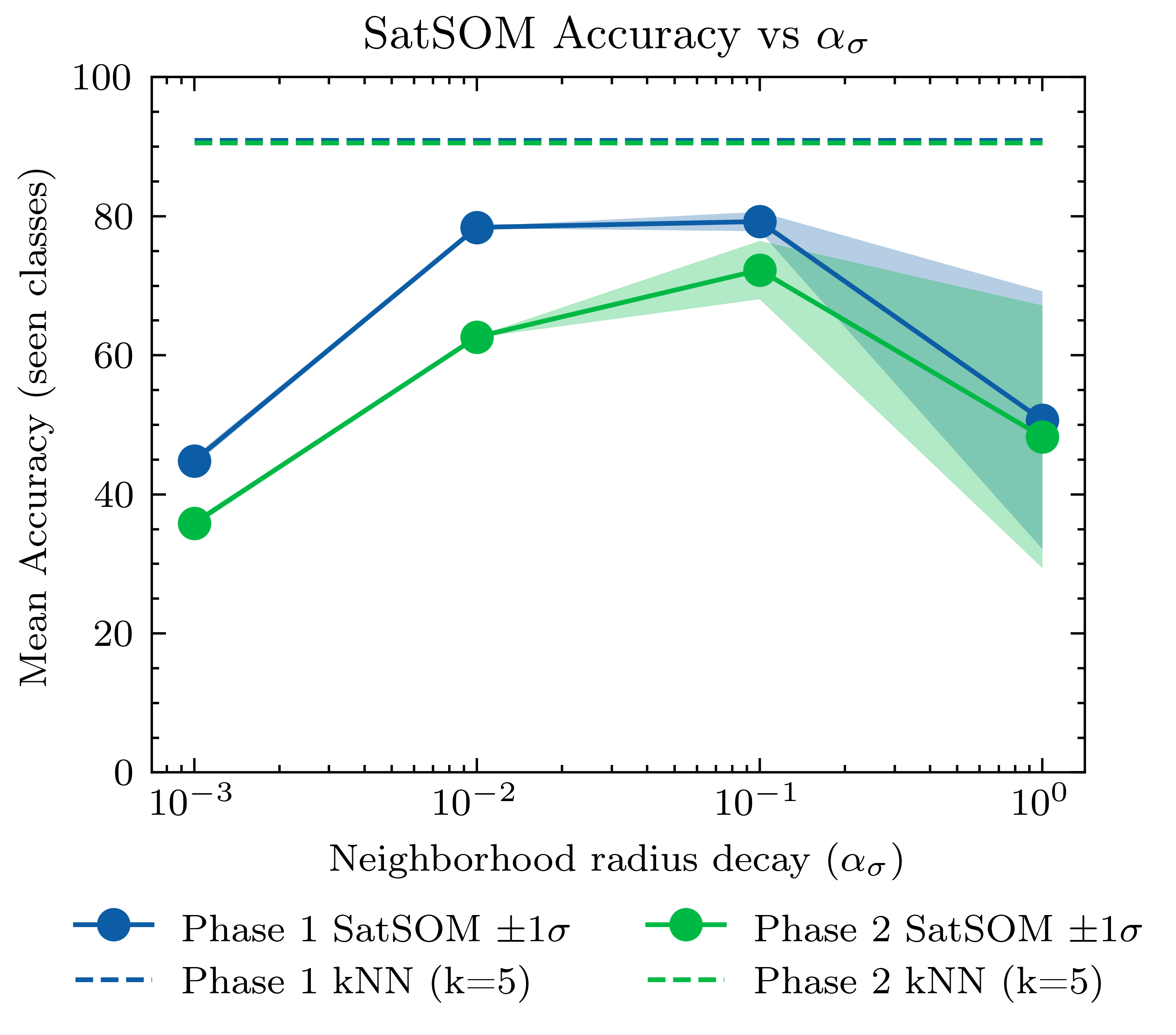}
\caption{Mean SatSOM accuracy as a function of the $\alpha_\sigma$ hyperparameter. Values near the beginning of the range produce unsatisfactory results, while higher values have a high standard deviation.}\label{fig:hyper_sigma}
\end{figure}

\begin{figure}
\centering
\includegraphics[width=\columnwidth]{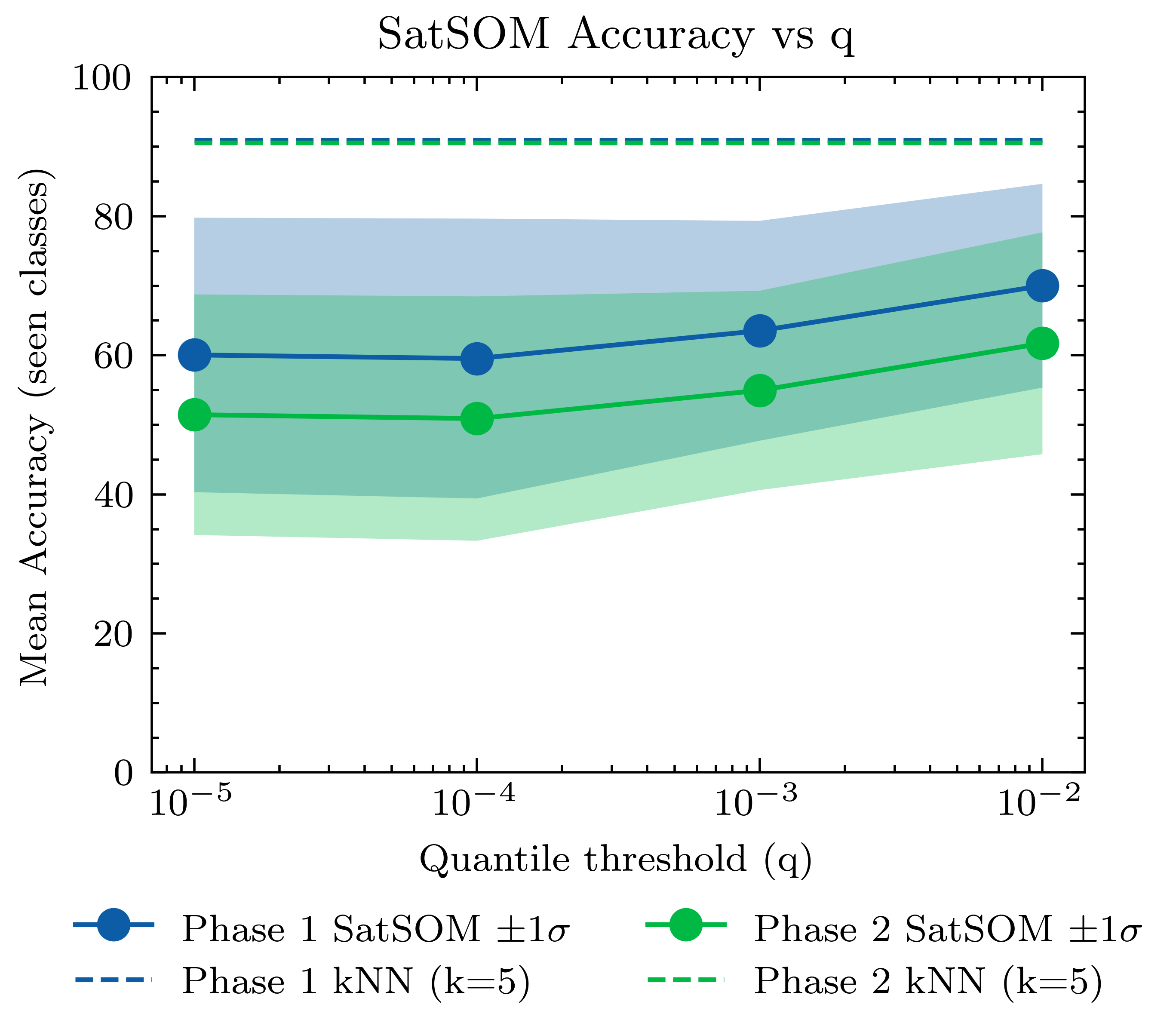}
\caption{Mean SatSOM accuracy as a function of the q hyperparameter. All values within the tested range yield satisfactory results.}\label{fig:hyper_q}
\end{figure}

To reduce computational cost during the hyperparameter search, we limited training to two phases. The first phase included classes 0, 1, 2, 4, 5, 6, 7, and 8 simultaneously, while the second phase comprised classes 3 and 9. SatSOM was trained 10 times for each hyperparameter configuration, and the results were subsequently aggregated. The presence of some evidently suboptimal parameter combinations led to poor performance in certain models, which affected the aggregated average accuracy and increased standard deviation. This effect is visible in the results presented in Figures~\ref{fig:hyper_n}, \ref{fig:hyper_q}, and \ref{fig:hyper_sigma}.

\begin{table}
\centering
\label{tab:satsom_phases}
\begin{tabular}{@{}ll@{}}
\toprule
\textbf{Phase 1} & \textbf{Phase 2} \\
\midrule
0, 1, 2, 4, 5, 6, 7, 8 & 3, 9 \\
\bottomrule
\end{tabular}
\caption{Classes used in the test}
\end{table}

As shown in Figure~\ref{fig:hyper_n}, and according to all our other tests, the grid size ($n$) has no observable effect on the model accuracy as long as it is large enough. It is, as expected, a primary determinant of the model's maximum capacity.

We found that the decay of the neighborhood radius ($\alpha_\sigma$) between 0.01 and 0.1 is suitable for most applications (Figure~\ref{fig:hyper_sigma}). Values outside of that range tend to produce unstable results or be clearly ineffective.

As shown in Figure~\ref{fig:hyper_q}, most of the quantile threshold values ($q$) worked relatively well, with a slight advantage on values closer to 0.01.

Although we do not have formal evidence, our empirical observations during testing suggest that reducing the initial neighborhood radius ($\sigma_0$) improves knowledge retention, albeit at the cost of requiring more input data. In contrast, increasing $\sigma_0$ accelerates training, but tends to reduce knowledge retention. Based on these findings, we recommend using a lower value of $\sigma_0$ when retention is a priority. Typically $\sigma_0=\frac{n}{2}$ is used as a starting point.

\section{Extension: Adaptive Capacity} 

As demonstrated in previous sections, the information capacity of a SatSOM is fundamentally constrained by its lattice size, N. However, the optimal model capacity for a continuous stream of tasks is rarely known \textit{a priori}. To address this, we propose an adaptive extension of the architecture: the \textit{Growing SatSOM}. We hypothesize that a dynamic expansion mechanism allows the model to approximate the optimal value of N relative to the task complexity. For computational simplicity, we restrict grid expansion to discrete increments equal to the initial grid dimension, n, maintaining the two-dimensional topology. Newly added neurons are initialized randomly.

\subsection{Boundary-Driven Expansion}

To mitigate boundary effects, the grid expands whenever a BMU is located within the effective neighborhood radius of an edge. We define a trigger condition controlled by a scaling factor $\beta \in [0, 1]$:

\begin{equation}
d_\text{edge} \le \max(\beta\sigma_\text{BMU}, 1),
\end{equation}

where $d_\text{edge}$ denotes the distance from the BMU to the grid boundary. $\beta=1$ ensures the full neighborhood is preserved, while $\beta=0$ limits sensitivity to the outermost neurons.

\begin{figure}
\centering
\includegraphics[width=\columnwidth]{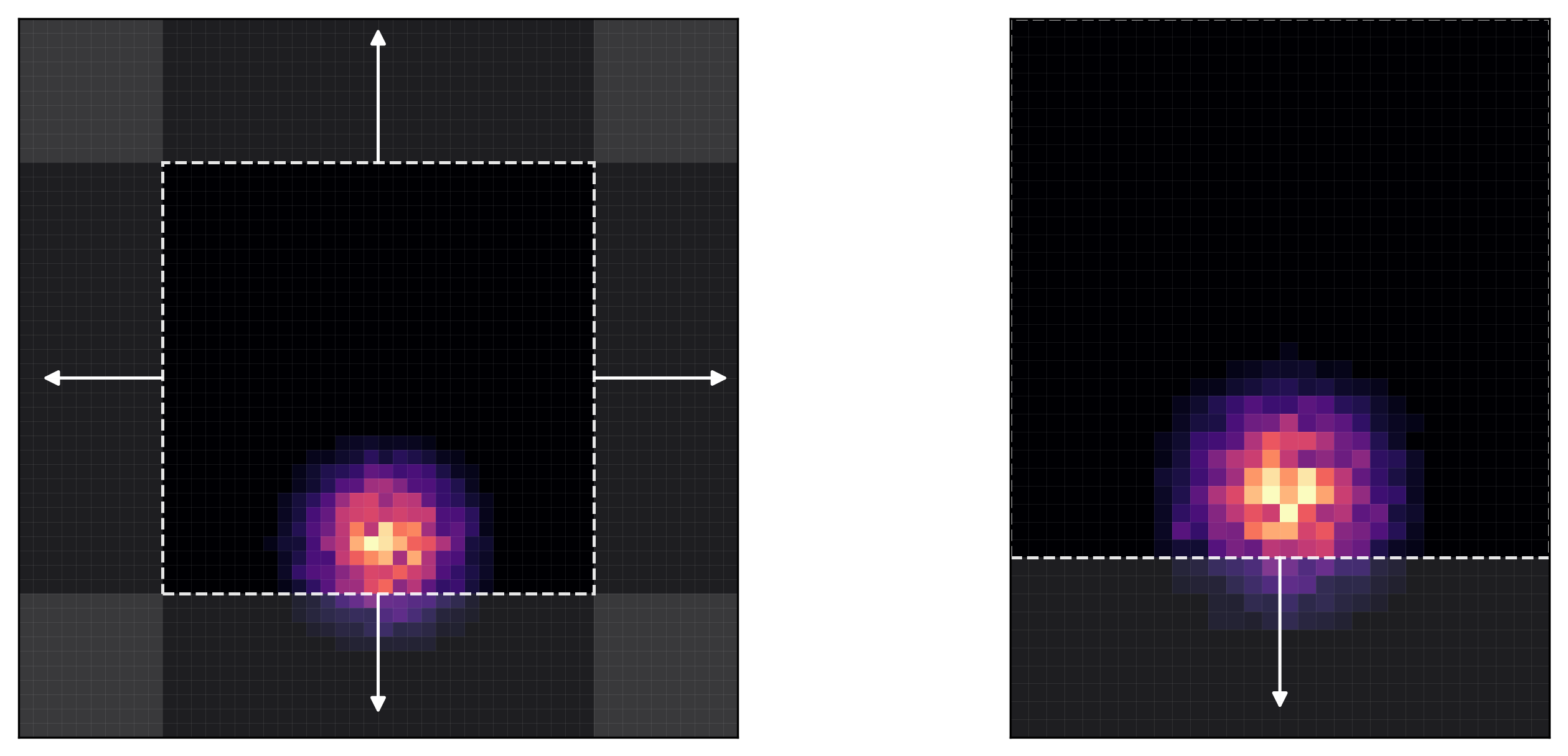}
\caption{Comparison of Symmetric (left) and Directional (right) grid growth schemes visualized on an example saturation map.}\label{fig:naive_vs_directional}
\end{figure}

We evaluate two expansion strategies: \textit{Symmetric}, which extends all grid dimensions uniformly, and \textit{Directional}, which extends only the boundary closest to the BMU, allowing for non-square topologies. Figure~\ref{fig:naive_vs_directional} illustrates these strategies.

\subsection{Grid Re-centering Strategy}

To optimize grid utilization and mitigate unnecessary expansion caused by spatial drift (cf. Figure~\ref{fig:saturations}), we introduce a re-centering mechanism. This process realigns the active region, defined as the set of neurons where $s_i > 0$, with the geometric center of the grid.

\begin{figure}
\centering
\includegraphics[width=\columnwidth]{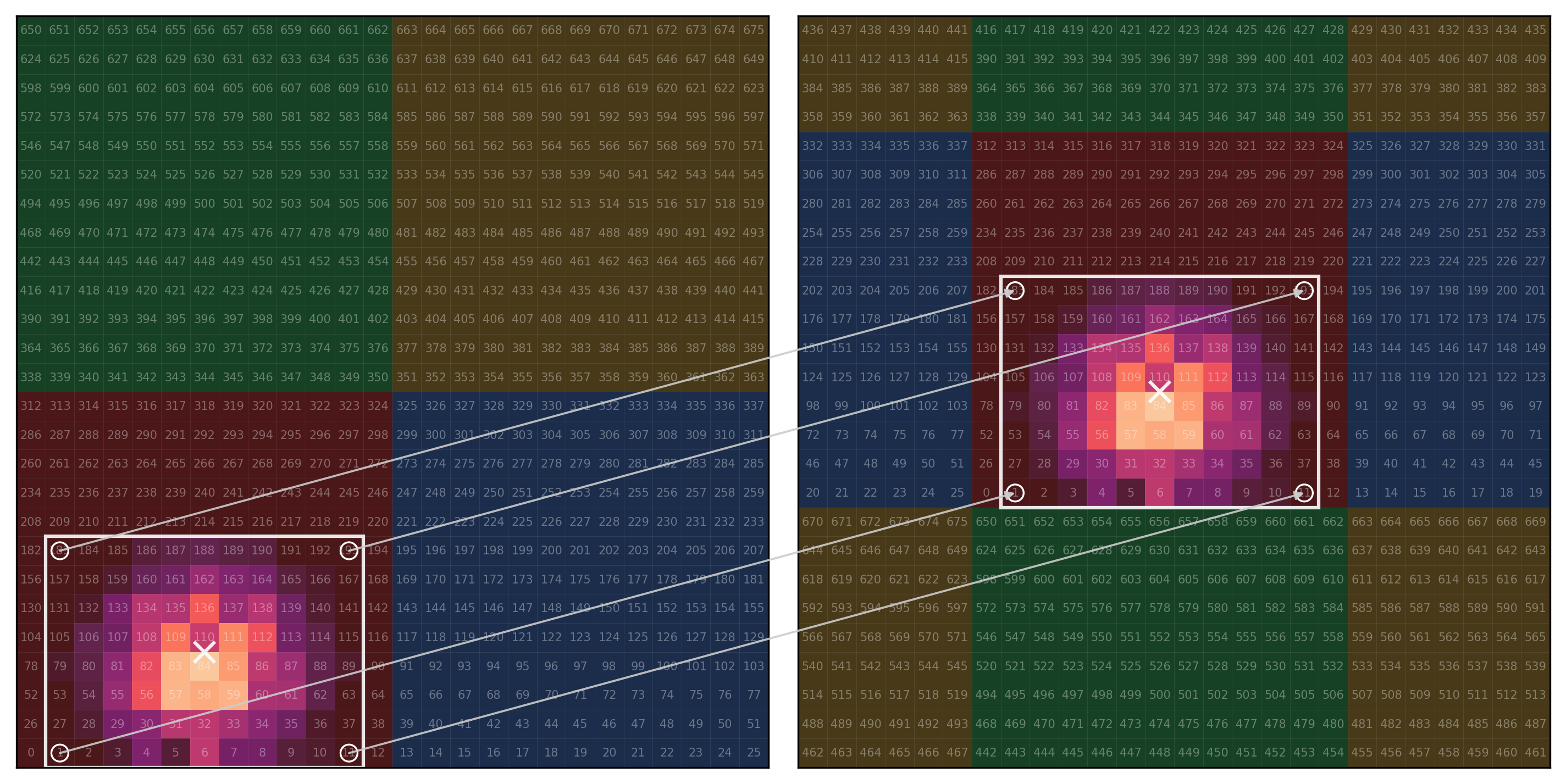}
\caption{Visualization of the toroidal shift operation on a saturation map. Neurons are numerically indexed, and quadrants are color-coded to trace the spatial transformation.}\label{fig:toroidal_rotation}
\end{figure}

The algorithm computes the axis-aligned bounding box of the active neurons and translates the entire grid using a toroidal shift (wrap-around) to center this bounding box. Since neurons with zero saturation do not contribute to the model output (Equation~\ref{eq:epsilon}), this transformation leaves the model functionally equivalent, with no change in classification accuracy. The operation is visualized in Figure~\ref{fig:toroidal_rotation}.

\subsection{Validation of the Dynamic Mechanism}

We evaluate the dynamic extension on CIFAR-10 (ResNet18 backbone) using the hyperparameters in Table~\ref{tab:hyperparams}, modified with $n=10$ and $\sigma_0=3$ to facilitate growth.

\begin{table}
\centering
\begin{tabular}{@{}ll@{}}
\toprule
\textbf{Parameter} & \textbf{Values} \\
\midrule
Scheme & Symmetric, Directional \\
Centering & TRUE, FALSE \\
$\beta$ & 0, 0.1, 0.2, ..., 0.9, 1 \\
$n$ & 10 \\
$\sigma_0$ & 3 \\
\bottomrule
\end{tabular}
\caption{Hyperparameters of the growing SatSOM experiment.}\label{tab:growing_parameters}
\end{table}

\begin{figure}
\centering
\includegraphics[width=0.9\columnwidth]{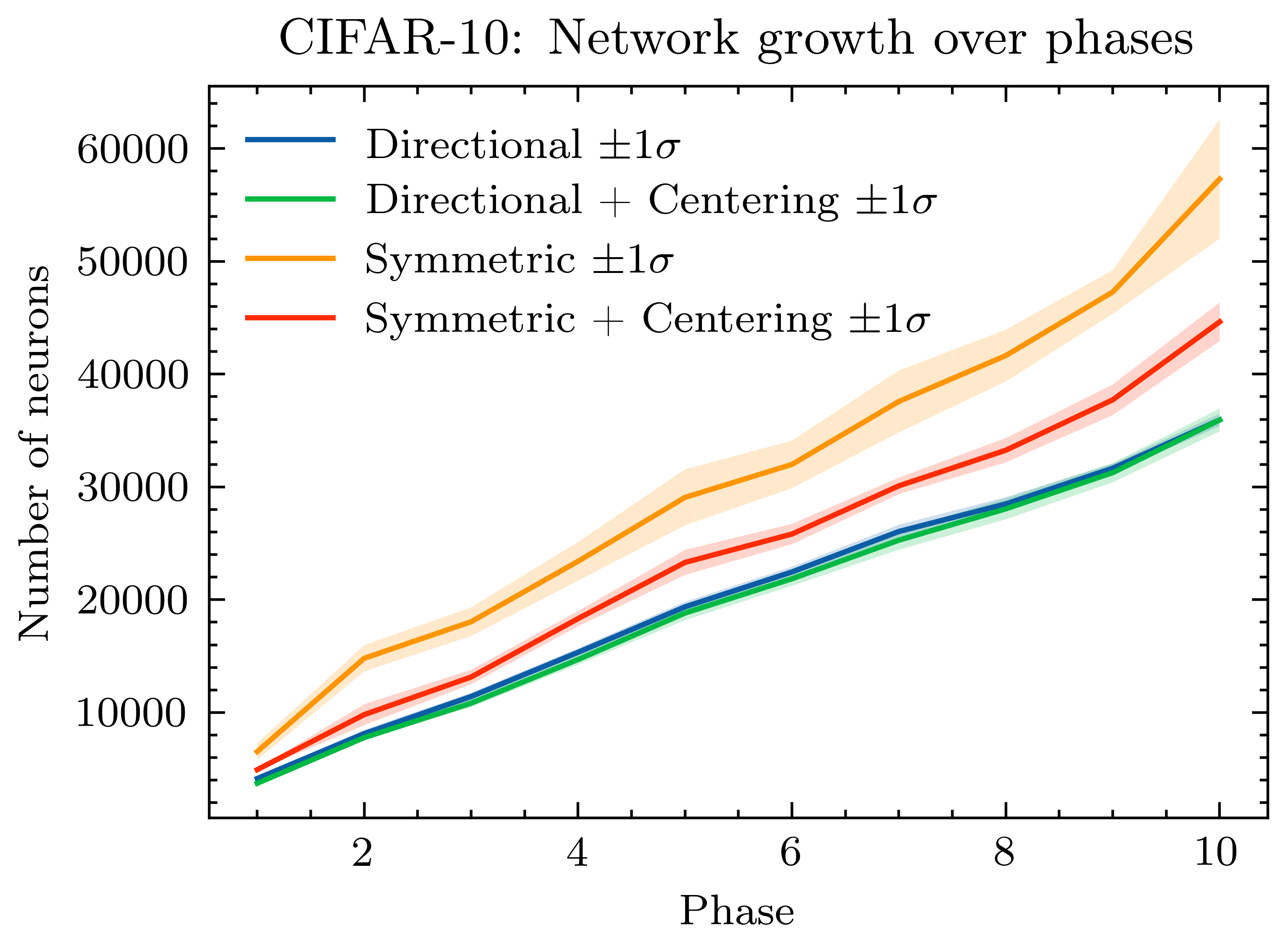}
\caption{Number of neurons ($N$) of a growing SatSOM through different phases of the class-incremental CIFAR-10 experiment. Aggregated by method hyperparameters.}\label{fig:overall_size}
\end{figure}

\begin{figure}
\centering
\includegraphics[width=0.9\columnwidth]{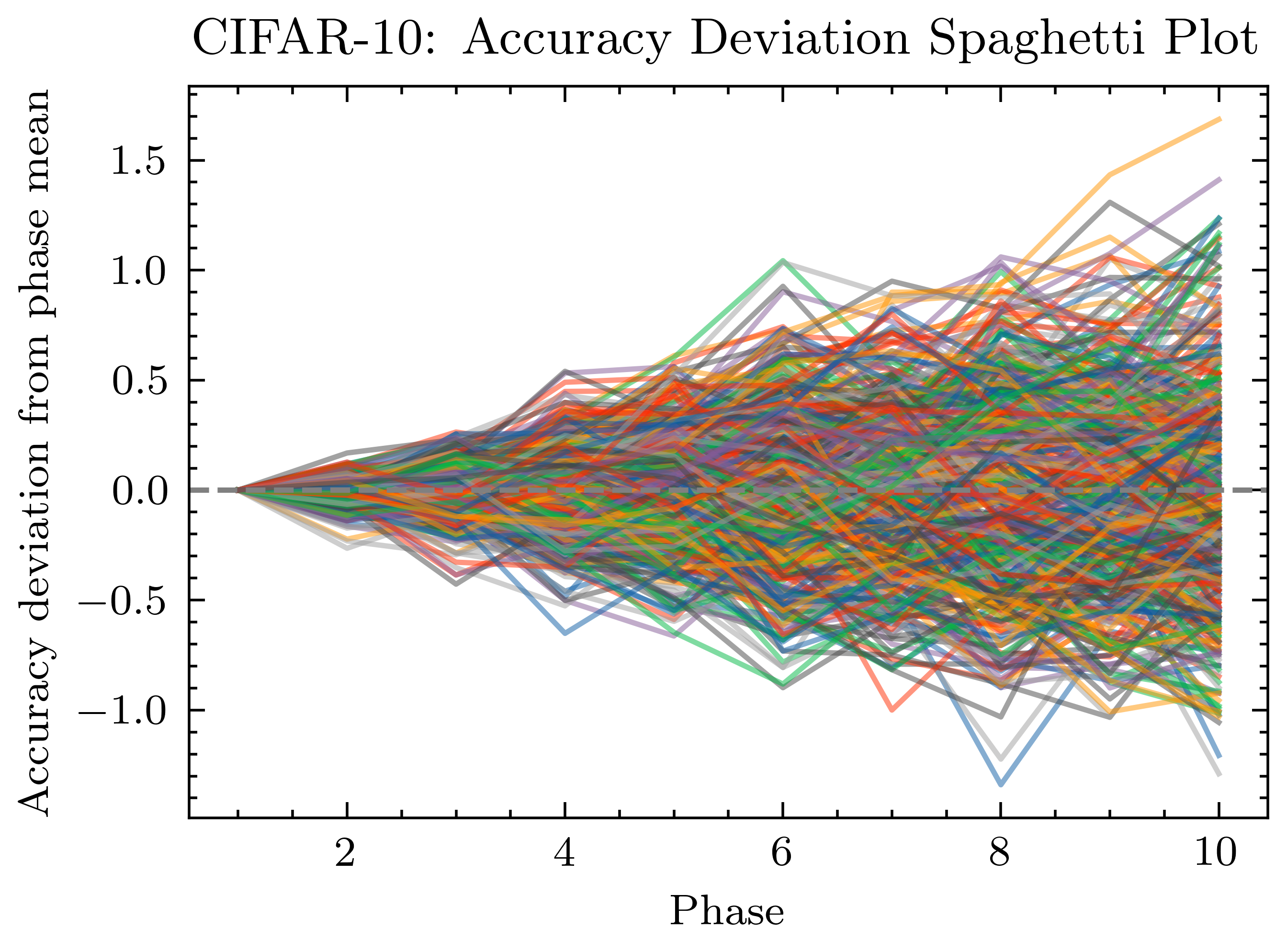}
\caption{Deviation from the mean accuracy of all of t he CIFAR-10 experiment runs (400 trials) through the phases. The deviation is not significant.}\label{fig:accuracy_spaghetti}
\end{figure}

Results indicate that the Directional scheme consistently produces more compact grids than the Symmetric baseline (Figure~\ref{fig:overall_size}). Centering further reduces grid size, particularly for the Symmetric scheme. As classification accuracy remains stable across all configurations (Figure~\ref{fig:accuracy_spaghetti}), we identify Directional growth with Centering as the optimal trade-off between storage efficiency and performance.

\begin{figure}
\centering
\includegraphics[width=0.9\columnwidth]{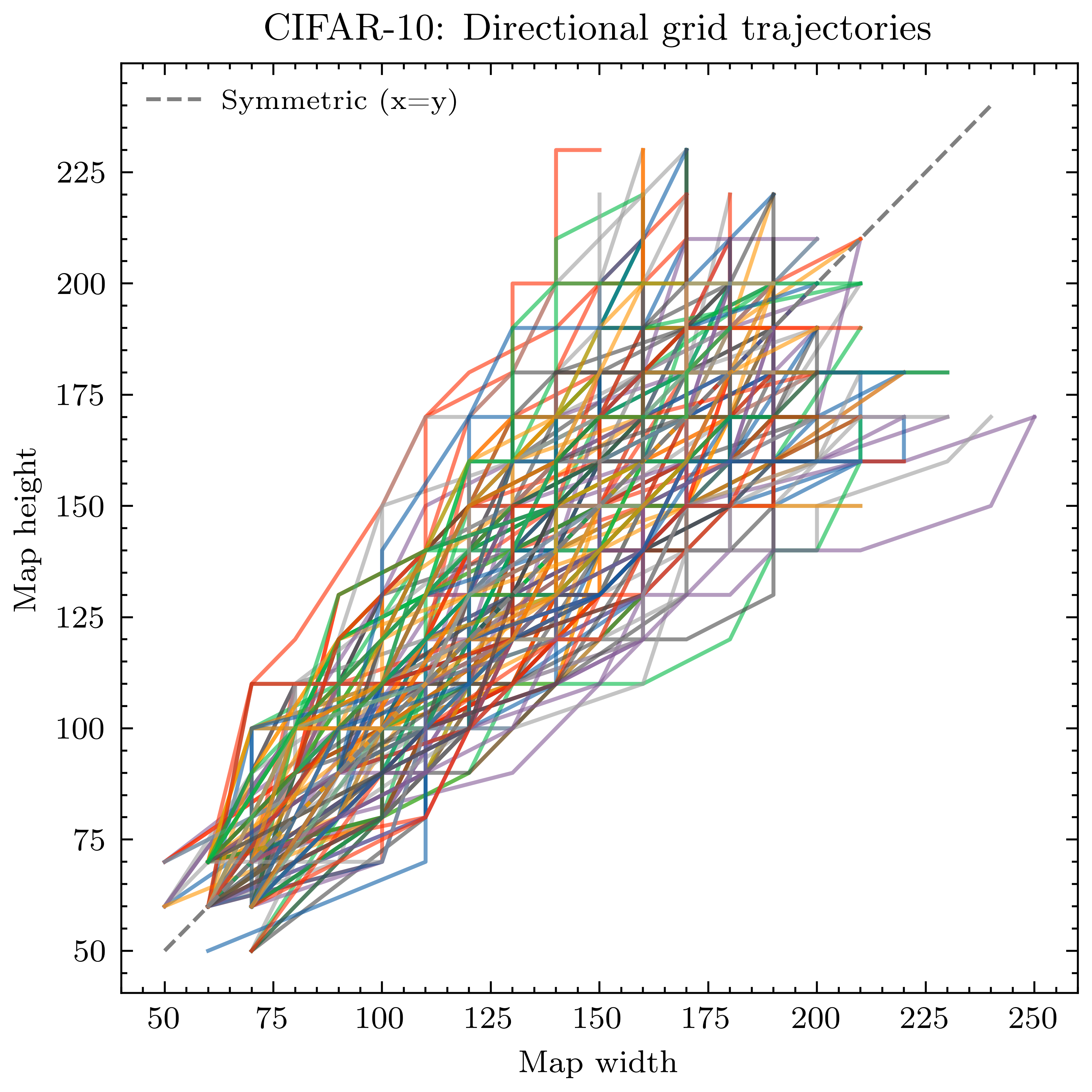}
\caption{Grid shape with the Directional scheme of all of the CIFAR-10 experiment runs (400 trials) through the phases. Symmetric square grid shape is marked as the dashed ($x=y$) line.}\label{fig:grid_shape}
\end{figure}

\begin{figure}
\centering
\includegraphics[width=0.9\columnwidth]{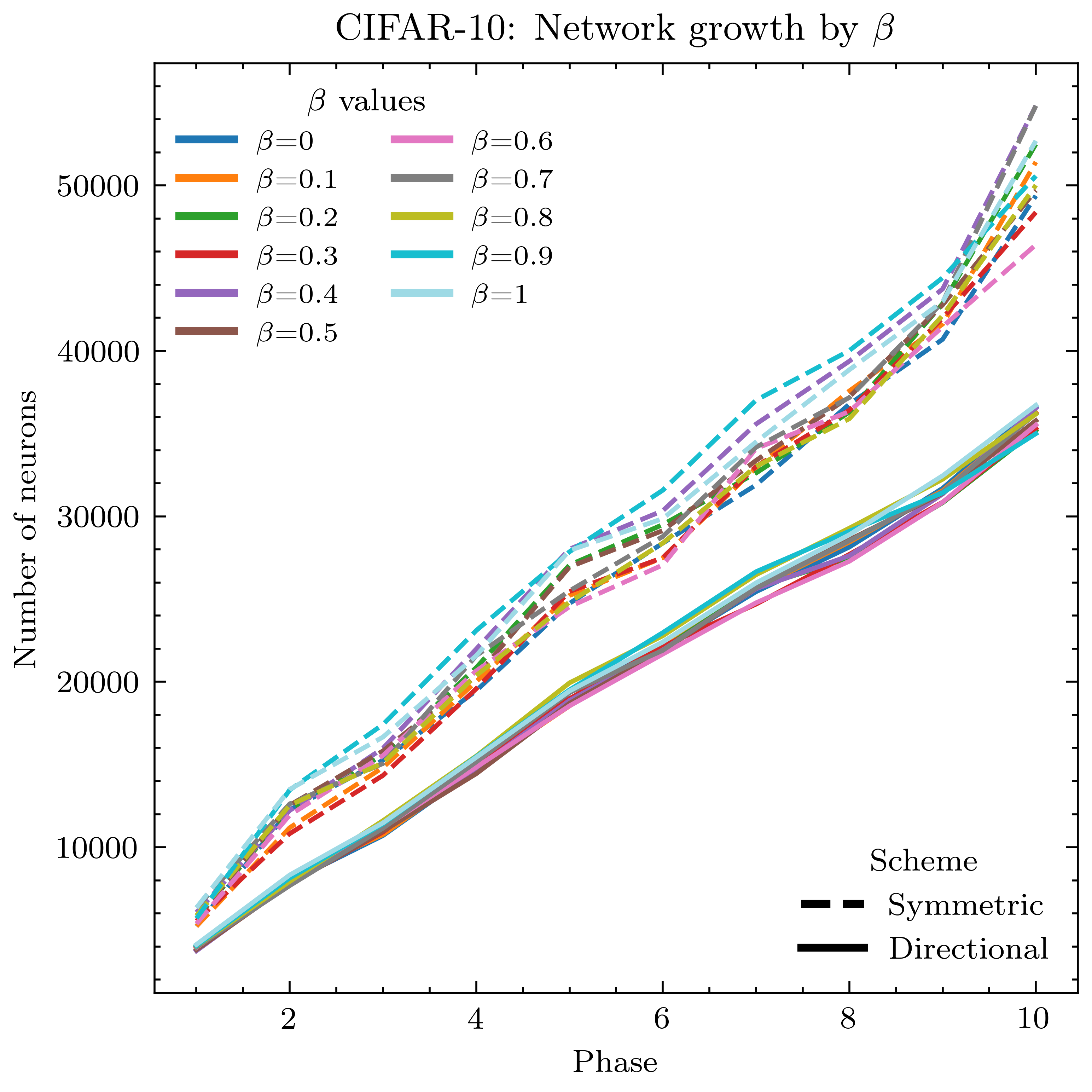}
\caption{Growing SatSOM size ($N$) aggregated according to the $\beta$ hyperparameter. No significant pattern is visible.}\label{fig:beta_plot}
\end{figure}

Regarding topology, the Directional scheme maintains a near-square aspect ratio throughout the training (Figure~\ref{fig:grid_shape}), with no significant distortion caused by Centering. Additionally, the sensitivity parameter $\beta$ showed negligible impact on these results (Figure~\ref{fig:beta_plot}).

\section{Conclusion}

In this paper, we introduced SatSOM, a novel extension of the SOM algorithm designed to mitigate catastrophic forgetting in continual learning scenarios. The core innovation of SatSOM is a saturation mechanism that modulates the learning rate and neighborhood radius based on individual neuronal training history. As a neuron accumulates knowledge, its plasticity diminishes, effectively stabilizing its weights. This dynamic directs new information toward unsaturated neurons, thereby minimizing interference with previously learned representations. Furthermore, the local nature of the model facilitates the removal of obsolete data through the selective re-initialization of prototypes.

Our empirical results demonstrate that SatSOM significantly outperforms other SOM-based methods in continual learning tasks. It yields performance competitive with DSDM, particularly on datasets with lower dimensionality, and exhibits retention capabilities approximating those of a kNN baseline. Crucially, ablation studies confirm that the saturation mechanism is the primary driver of these improvements, underscoring its efficacy in managing the stability-plasticity trade-off.

This research contributes to the development of continual learning systems that are efficient, stable, and compatible with strict online constraints. Unlike state-of-the-art approaches that often require external memory buffers, explicit task boundaries, or generative rehearsal, SatSOM offers a lightweight, interpretable alternative. Its fixed memory footprint makes it particularly promising for resource-constrained applications.

To address scenarios where task complexity is unknown \textit{a priori}, we also introduced a dynamic variant: the Growing SatSOM. This extension demonstrates that simple, boundary-driven expansion rules allow the model to adapt its capacity on the fly. By combining directional growth with a grid re-centering strategy, the model approximates the optimal network size for a given task without compromising the storage efficiency or stability of the core algorithm.

The primary limitation of the current approach lies in the inherent trade-off between local specificity and global abstraction. As the effective neighborhood radius decays, the network becomes increasingly \textit{localist}—a phenomenon also observed in CSOM~\cite{CSOM}. While this behavior enhances retention, it constrains the model's ability to capture high-level abstractions that span large regions of the data manifold. Future work will address this by exploring hierarchical architectures, similar to those proposed in~\cite{ghsom}, to enable global abstraction while maintaining continual learning capabilities at the local level.

Several other promising research directions emerge from this study. First, the saturation principle could be adapted to deep neural network architectures, potentially through layer-wise or neuron-specific plasticity control to mitigate forgetting in deep representations. Similarly, the mechanism aligns well with neuromorphic paradigms; self-organizing variants of Operational Neural Networks, such as those proposed by Zhang et al~\cite{SOONN}, could integrate saturation-based plasticity with high biological fidelity.

Beyond architectural modifications, SatSOM shows potential for integration into hybrid systems, acting as a guide for replay mechanisms~\cite{memory_efficient_replay}. Finally, exploring the deployment of SatSOM on resource-constrained hardware would validate its utility for on-device learning in robotics and embedded environments.

Ultimately, SatSOM offers a biologically grounded and visually interpretable framework for continual learning. Our findings confirm that the adaptive modulation of plasticity is a potent mechanism for mitigating catastrophic forgetting. This work establishes a robust foundation for scaling these principles to complex architectures and deployment in open-ended, dynamic environments.

\section{Acknowledgements}

We thank Dr. Leszek Grzanka for his support in running the experiments.

This research was supported by the funds assigned to AGH University by the Polish Ministry of Education and Science.

We gratefully acknowledge Polish high-performance computing infrastructure PLGrid (HPC Center: ACK Cyfronet AGH) for providing computer facilities and support within computational grant no. PLG/2025/018713.

\section{Compliance with Ethical Standards}

This research adheres to all applicable ethical standards and guidelines. All datasets used in this work are publicly
available and widely used in the research community for benchmarking machine learning models.
The authors declare that they have no conflict of interest.

\bibliography{bibliography}
\end{document}